\documentclass[10pt,journal,cspaper,compsoc]{IEEEtran}

\usepackage{lmodern}
\usepackage{times}
\usepackage{graphicx}
\usepackage{amsmath}
\usepackage{amssymb}
\usepackage{cite}
\usepackage{fancyhdr}
\usepackage{color,colortbl}
\usepackage{multirow}
\usepackage{tabularx}
\usepackage{color}
\usepackage[table]{xcolor}
\usepackage{overpic}
\usepackage{wrapfig,lipsum,booktabs,subfigure}
\usepackage{url}
\usepackage{makecell}
\usepackage{algorithm}
\usepackage{algpseudocode}
\usepackage{amsmath}
\usepackage{graphics}
\usepackage{ragged2e}
\usepackage{silence}
\graphicspath{{./fig/}}

\usepackage[pagebackref=true,breaklinks=true,colorlinks,bookmarks=false]{hyperref}

\renewcommand{\eqref}[1]{Eqn.~(\ref{#1})}
\renewcommand{\algref}[1]{Algorithm~(\ref{#1})}

\newcommand{\myPara}[1]{\vspace{.2in}\noindent \textbf{#1}}

\newcommand{\tbf}[1]{\textbf{#1}}
\newcommand{\figref}[1]{Fig.~\ref{#1}}%
\newcommand{\tabref}[1]{Tab.~\ref{#1}}%
\newcommand{\secref}[1]{Sec.~\ref{#1}}
\renewcommand{\eqref}[1]{Equ.~(\ref{#1})}

\def\ie{\emph{i.e.,~}}
\def\eg{\emph{e.g.,~}}
\def\etc{\emph{etc}}
\def\etal{{\em et al.}}

\begin{document}


\title{RF-Next: Efficient Receptive Field Search for Convolutional Neural Networks}

\author{Shanghua Gao \qquad Zhong-Yu Li  \qquad Qi Han
	\qquad Ming-Ming Cheng  \qquad Liang Wang
	\IEEEcompsocitemizethanks{
		\IEEEcompsocthanksitem S. Gao, Z.Y Li, Q, Han, and M.M. Cheng are with
		the TMCC, CS, Nankai University,
		Tianjin 300350, China.
		\IEEEcompsocthanksitem L Wang is the with National Laboratory of Pattern Recognition.
		\IEEEcompsocthanksitem M.M. Cheng (cmm@nankai.edu.cn)
		is the corresponding author.
		\IEEEcompsocthanksitem A preliminary version of this work has been presented in the CVPR 2021~\cite{gao2021global2local}.
	}
}

\markboth{IEEE Transactions on Pattern Analysis and Machine Intelligence}
{Gao \MakeLowercase{\textit{et al.}}:
	RF-Next: Efficient Receptive Field Search for Convolutional Neural Networks}

\IEEEtitleabstractindextext{
	\begin{abstract}
		\justifying
		Temporal/spatial receptive fields of models play an important role in sequential/spatial tasks.
		Large receptive fields facilitate long-term relations,
		while small receptive fields help to capture the local details.
		Existing methods construct models with hand-designed receptive fields in layers.
		Can we effectively search for receptive field combinations
		to replace hand-designed patterns?
		To answer this question,
		we propose to find better receptive field combinations
		through a global-to-local search scheme.
		Our search scheme exploits both global search to find the coarse combinations
		and local search to
		get the refined receptive field combinations further.
		The global search finds possible coarse combinations
		other than human-designed patterns.
		On top of the global search,
		we propose an expectation-guided iterative local search scheme to
		refine combinations effectively.
		Our RF-Next models, plugging receptive field search to various models, boost the performance on many tasks, \eg temporal action segmentation,
		object detection, instance segmentation, and speech synthesis.
		The source code is publicly available on \url{http://mmcheng.net/rfnext}.
	\end{abstract}
	}

\maketitle
\IEEEdisplaynontitleabstractindextext
\IEEEpeerreviewmaketitle

\IEEEraisesectionheading{\section{Introduction}\label{sec:introduction}}

Due to the strong representation ability,
convolutional neural networks (CNN) have been widely used
in spatially visual recognition tasks, \eg object detection~\cite{fastrrcnn},
saliency detection~\cite{21PAMI-Sal100K},
instance segmentation~\cite{He_2017}, and semantic segmentation~\cite{chen2018deeplab,gao2021large},
as well as sequential perception tasks,
\eg temporal action segmentation~\cite{MS-TCN-PAMI20,chen2020action}, and speech synthesis~\cite{PingPZ020,li2019neural}.
CNN processes short/long-term features by stacking convolutional filters with different receptive fields.
Spatial convolutional networks (SCN) for visual recognition tasks
process local and global features
to represent the texture and semantic information.
Temporal convolutional networks (TCN)~\cite{lea2017temporal,farha2019ms,MS-TCN-PAMI20,wangboundary,fayyaz2020sct}
are widely adopted in sequential tasks with their ability
to capture both long-term and short-term information.
Appropriate receptive fields in layers are crucial for both SCN and TCN as
large receptive fields contribute to long-term dependencies
while small receptive fields benefit the local details.
State-of-the-art (SOTA) SCN~\cite{he2016deep,pami20Res2net,fastrrcnn,He_2017,gao2021rbn} and TCN~\cite{MS-TCN-PAMI20,chen2020action,wangboundary,li2020set,huang2020improving}
methods rely on human-designed receptive field combinations,
\ie dilation rate or pooling size in each layer,
to make the trade-off between capturing long and short term dependencies.
Questions have been raised: Are there other effective receptive field combinations
that perform comparable or better than hand-designed patterns?
Will the receptive field combinations vary among different datasets?
To answer those questions,
we propose to find the possible receptive field combinations in a
coarse-to-fine scheme through the global-to-local search.

\begin{figure}[!t]
	\centering
	\begin{overpic}[width=\linewidth]{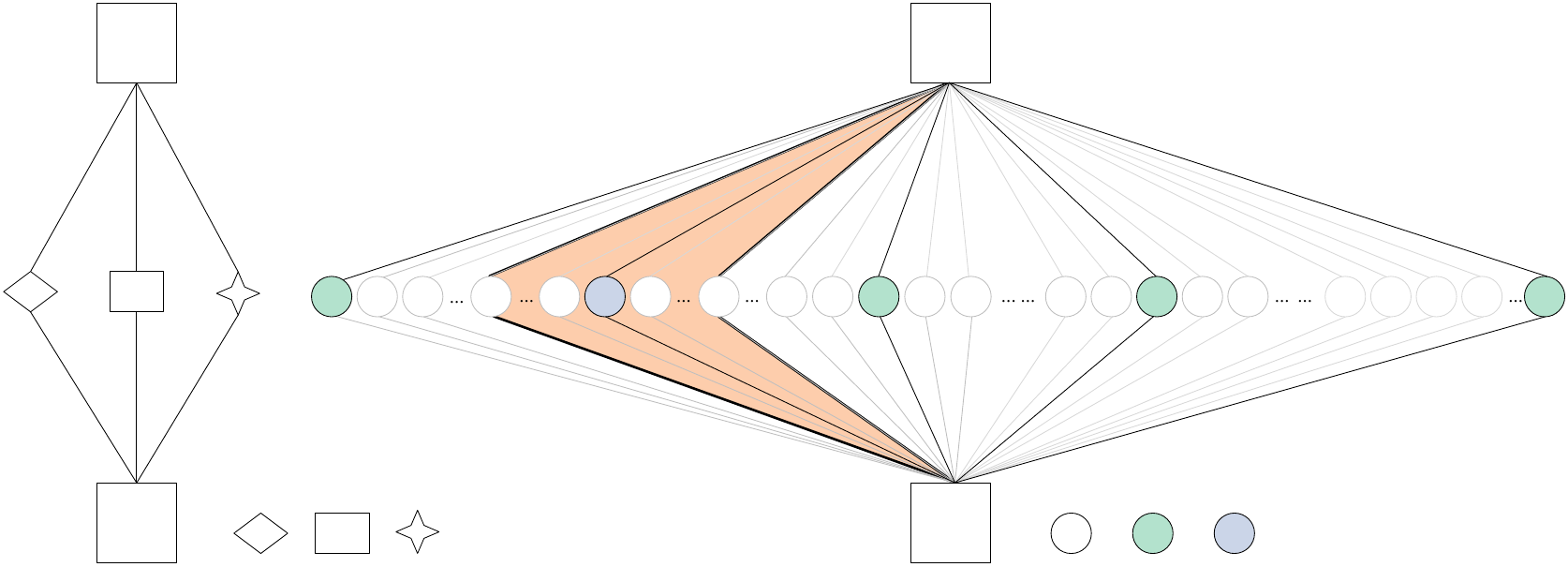}
		\put(30, 1){operators}
		\put(83, 1){dilations}
	\end{overpic}
	\caption{Search space comparison between searching for network architecture
		and receptive field combinations.
		Left: Network architecture search mostly searches for several operations
		with different functions.
		Right: The search space of receptive field combinations is huge.
		The white, green, blue nodes, and orange shade represent
		the dilation rate candidates,
		the sparse search space in global search,
		one of the global searched results,
		and the local search space, respectively.
	}\label{fig:tasks}
\end{figure}

As shown in~\figref{fig:tasks},
unlike the existing network architecture search spaces~\cite{liu2019darts,cai2018proxylessnas,howard2019searching}
that only contain several operation options within a layer,
the available search space of receptive field combinations could be huge.
Suppose a TCN/SCN has $L$ convolutional layers and $D$ possible receptive fields
in each layer, there are $D^L$ possible combinations,
\ie the MS-TCN~\cite{farha2019ms} for the long-sequence temporal action segmentation task,
consisting of 40 layers and 1024 possible receptive fields
in each layer,
has $1024^{40}$ possible receptive field combinations.
Directly applying network architecture searching algorithms
\cite{howard2019searching,liu2019auto,liu2019darts,xie2017genetic}
to such a huge search space is impractical.
For example, conventional reward-based searching methods
\cite{real2019regularized,liu2017hierarchical,xie2017genetic}
are unsuitable for CNN-based models with a huge search space.
The model training and performance evaluation of each possible combination are too costly.
Differentiable architecture searching methods (DARTS)
\cite{liu2019darts,cai2018proxylessnas,liu2019auto}
rely on shared big networks to save training time,
thus only supporting several operators within a layer due to
the constraint of model size.
Moreover, they are heavily dependent on the initial combination and fail to
find new combinations with a huge difference from the initial one.
While our goal is to explore effective receptive field combinations other than
human-designed patterns in the huge search space,
those algorithms are either too costly or cannot support the large search space.

To explore the effective receptive fields with a low cost,
we exploit both a genetic-based global search to find
the coarse receptive field combinations
and an expectation-guided iterative (EGI) local search to get
the refined combinations.
Specifically, we follow common settings in many existing methods
\cite{farha2019ms,chen2018deeplab,PrengerVC19,oord2016wavenet}
to use dilation rates to determine layers' receptive fields.
A genetic-based global search scheme is proposed to find coarse combinations
within a sparsely sampled search space at an affordable cost.
The global search discovers various combinations that achieve
even better performance
than human designings but have completely different patterns.
Based on the global-searched coarse combinations,
we propose the local search to determine fine-grained dilation rates.
In local search, a convolutional weight-sharing scheme enforces learned dilation coefficients
to approximate the probability mass distribution for calculating the expectation of dilation rates.
The expectation-guided searching transfers the discrete
dilation rates into a distribution,
allowing fine-grained dilation rates search.
With an iterative searching process,
the local search gradually finds more effective fine-grained receptive field
combinations with a low cost.
Models enhanced by our proposed global-to-local search scheme, namely RF-Next models,
surpass human-designed structures with impressive performance gain on many tasks.
In summary, we make two major contributions:
\begin{itemize}
	\item The expectation-guided iterative local search scheme enables
	      searching fine-grained receptive field combinations
	      in the dense search space.
	\item The global-to-local search discovers effective receptive field
	      combinations with better performance than hand-designed patterns.
\end{itemize}

The conference version \cite{gao2021global2local}
mainly explores the proposed receptive field searching scheme
on the temporal action segmentation task.
In this work, we improve and give more analysis of our proposed searching scheme,
\ie parallel receptive fields for multi-scale enhancement,
searching cost analysis, generalization to different tasks, observations of receptive fields.
We generalize our proposed receptive field searching method
to multiple tasks on both temporal and spatial dimensions,
\ie speech synthesis, sequence modeling,
instance segmentation, object detection,
and semantic segmentation.
To meet the multi-scale requirements of some tasks,
we take advantage of the proposed expectation-guide search scheme
to enable the searched structure with parallel multiple
receptive fields and shared convolutional weights.
We give observations on the receptive field requirements of multiple tasks
based on the searched results:
1) Proper receptive fields of CNN are beneficial to many tasks.
2) The receptive field requirements of different parts of the network are quite different.
\section{Related Work}

\subsection{Receptive Field in Networks}
The effect of receptive fields has been widely studied~\cite{luo2016understanding,loog2017supervised,wang2019receptive,chen2017rethinking,chen2019drop,tomen2021deep}.
Although the theoretical receptive field could be huge,
\cite{luo2016understanding} shows that the
effective receptive field occupies a fraction of the theoretical receptive field.
\cite{loog2017supervised} assumes that useful predictive information for a pixel comes
from nearby locations rather than far pixels
and proposes to gradually suppress the distant pixel values with the scale-sensitive regularization.
\cite{seif2018large} enlarges the receptive field to improve the performance of the image super-resolution task.
\cite{wang2019receptive} observes that model depth must be congruent with the receptive field size for the image super-resolution task.
\cite{tomen2021deep} forms continuous receptive fields with the help of Gaussian scale-space representation.
Parallel receptive fields are proposed to enable a more flexible receptive field within a layer~\cite{szegedy2015going,chen2017rethinking,chen2019drop}.
Inception net series~\cite{szegedy2015going,szegedy2016rethinking,szegedy2017inception}
explores the parallel asymmetric convolutions with different receptive fields to enhance the model representation ability.
Atrous spatial pyramid modules~\cite{chen2018deeplab,chen2017rethinking} prove the effectiveness of parallel receptive fields in semantic segmentation.
OctConv~\cite{chen2019drop} decomposes the convolution to process two feature scales simultaneously.
Some works model the connections of all positions to form arbitrary receptive fields theoretically~\cite{vaswani2017attention,wang2018non,chen2019graph}.
Multi-head attention is presented by~\cite{vaswani2017attention}
to model the relation between every two pixels with attention mechanisms.
Similarly, \cite{wang2018non} utilizes the non-local operation to aggregate features
at all positions.
A graph-based global reasoning module~\cite{chen2019graph} is proposed to capture
relations between arbitrary regions.
Despite these methods exploring receptive fields' great potential,
choosing effective receptive fields for different tasks is still an open question.

\subsection{Sequential Tasks}
Sequential tasks process data in the form of sequences, \eg video stream and audio stream.
As the sequence length could have a large variance for sequential tasks,
models with a proper range of effective receptive fields are needed.
In this work,
we mainly tackle two sequential tasks with long sequences of data, \ie temporal action segmentation and speech synthesis,
which represent video and audio dimensions, respectively.

\subsubsection{Temporal Action Segmentation}
Temporal action recognition segments the action of each video frame,
playing an important role in computer vision applications
such as clips tagging~\cite{soleymani2011multimodal},
video surveillance~\cite{collins2000introduction, collins2000system},
and anomaly detection~\cite{saligrama2012video}.
While conventional works~\cite{simonyan2014two,feichtenhofer2017spatiotemporal,carreira2017quo,feichtenhofer2019slowfast}
have continuously refreshed the recognition performance of short trimmed videos
containing a single activity,
segmenting each frame densely in long untrimmed videos remains
challenging as those videos contain many activities
with different temporal lengths.
In this work, we do our major experiments on the temporal action segmentation task.
Therefore, we give a thorough introduction to related works for the temporal action segmentation task.

Many approaches have been proposed for modeling dependencies for
temporal action segmentation.
Early works~\cite{fathi2013modeling,fathi2011understanding,fathi2011learning}
mostly model the changing state of appearance and actions with sliding
windows~\cite{rohrbach2012database,karaman2014fast,bhattacharya2014recognition}.
Thus they mainly focus on short-term dependencies.
Capturing both short-term and long-term dependencies then gradually becomes
the focus of temporal action segmentation.

\myPara{Sequential Model.}
Sequential models capture long-short term dependencies in an iterative form.
Vo and Bobick \cite{vo2014stochastic} apply the Bayes network to
segment actions represented by a stochastic context-free grammar.
Tang \etal~\cite{tang2012learning} use a hidden Markov model
to model transitions between states and durations.
Later, hidden Markov models are combined with
context-free grammar~\cite{kuehne2016end},
Gaussian mixture model~\cite{kuehne2017weakly},
and recurrent networks~\cite{richard2017weakly,kuehne2018hybrid}
to model long-term action dependencies.
Cheng~\etal~\cite{cheng2014temporal} apply the sequence memorizer
to capture long-range dependencies in visual words learned from the video.
However, these sequential models are inflexible in parallelly modeling
long-term dependencies and usually suffer from
information forgetting \cite{farha2019ms,MS-TCN-PAMI20}.

\myPara{Multi-stream Architecture.}
Some researchers
\cite{richard2016temporal,singh2016multi,singh2016first,ding2017tricornet}
utilize multi-stream models to model dependencies
from long and short term.
Richard and Gall employ \cite{richard2016temporal} dynamic programming
to inference models composed of length model,
language model, and action classifier.
Singh~\etal~\cite{singh2016multi} learn short video chunks representation
with a two-stream network and pass these chunks to a bi-directional network
to predict temporal action segmentation results sequentially.
A three-stream architecture is proposed in \cite{singh2016first},
which contains egocentric cues, spatial and temporal streams.
Tricornet~\cite{ding2017tricornet} utilizes a hybrid temporal convolutional
and recurrent network to capture local motion and
memorize long-term action dependencies.
CoupledGAN~\cite{gammulle2019coupled} uses a GAN model to utilize
multi-modal data to better model human actions' evolution.
Capturing long-short term information with multiple streams increases
computational redundancy.

\myPara{Temporal Convolutional Network.}
Recently, temporal convolutional networks (TCN) have been introduced to model
dependencies of different ranges within a unified structure by adjusting receptive fields and can process long videos in parallel.
Lea~\etal~\cite{lea2017temporal} propose the encoder-decoder style TCN
for the temporal action segmentation to capture long-range temporal patterns
and apply the dilated convolution to enlarge the receptive field.
TDRN~\cite{lei2018temporal} further introduces the deformable convolution
to process the full-resolution residual stream and low-resolution pooled stream.
MS-TCN~\cite{farha2019ms,MS-TCN-PAMI20} utilizes multi-stage dilated TCNs with
hand-designed dilation rate combinations to capture information from
various temporal receptive fields.
However, the adjustment of receptive fields still relies on human design,
which may not be appropriate.
Our proposed efficient receptive field combinations searching scheme
can automatically discover more efficient structures,
improving these TCN based methods.

\myPara{Complementary Techniques.}
Instead of capturing long-term and short-term information,
some works~\cite{ding2018weakly,wangboundary} further improve the
temporal action segmentation performance with boundary refinement.
Li~\etal~\cite{ding2018weakly} utilize an iterative training procedure
with transcript refinement and soft boundary assignment.
Wang~\etal~\cite{wangboundary} leverage semantic boundary information
to refine the prediction results.
Other researchers focus on temporal action segmentation under the
weakly supervised~\cite{kuehne2017weakly,richard2017weakly,ding2018weakly},
or unsupervised~\cite{sener2018unsupervised} settings.
These works still rely on the efficient TCN to model the action dependencies,
thus complementing the proposed method.

\subsubsection{Speech Synthesis}
\label{sec:tts_related}
Speech synthesis, also known as text to speech (TTS),
aims to synthesize human-like natural speech from text~\cite{dutoit1997introduction,taylor2009text},
which has thrived due to the strong feature representation ability of the neural networks
~\cite{PrengerVC19,KimLSKY19,PingPZ020,li2019neural}.
The large difference in modality and feature-length between text and speech
makes it hard to implement TTS in an end-to-end style~\cite{tan2021survey}.
Therefore, common approaches for speech synthesis
decompose this process into the following steps:
1) text-to-linguistic features transformation~\cite{panayotov2015librispeech,arik2017deep}.
2) transfer linguistic features or text to acoustic features~\cite{yoshimura1999simultaneous,ze2013statistical,Wang2017,li2019neural}.
3) generate the final speech waveform using the acoustic features~\cite{oord2016wavenet,PrengerVC19,KimLSKY19,PingPZ020}.
During the process, features are transformed from the short-length text sequence
to the long-length acoustic features and speech waveform,
\ie text with about 20 words results in a 5-second speech sequence of 80k sampling points.
We tackle the challenge of transferring acoustic features to speech waveform,
as this procedure requires a suitable receptive field to model the short/long-term
dependencies in the waveform~\cite{tan2021survey,PrengerVC19}.
Since we only focus on the receptive field of TTS models,
we refer readers to the survey~\cite{tan2021survey} for more details of speech synthesis.

\subsection{Spatial Tasks}
Unlike the sequential tasks that mostly process one-dimension sequences,
spatial tasks process images with two dimensions,
\ie height and width dimensions.
To extract features of objects of various sizes in the scene,
the model needs small receptive fields to detect small objects
and large receptive fields to cover large objects or capture surrounding context information~\cite{pami20Res2net}.
We mainly tackle two popular vision tasks
with the proposed receptive field search,
\ie object detection and instance segmentation.

\subsubsection{Object Detection}
Object detection aims to localize objects with bounding boxes
and assign categories accordingly~\cite{detectors,Sun_2021_CVPR,Xiong_2021_CVPR,Wang_2021_CVPR,Liang_2021_CVPR}.
Common object detection methods can be divided into single-stage~\cite{liu2016ssd,redmon2016you,law2018cornernet,Sun_2021_CVPR,Xiong_2021_CVPR} and two-stage~\cite{fastrrcnn,Cai_2019,chen2019hybrid} pipelines.
More details about object detection can be referred to related surveys~\cite{zou2019object,zaidi2021survey,liu2020deep}.
Single-stage detectors, \eg SSD~\cite{liu2016ssd}, YOLO~\cite{redmon2016you}, and CornerNet~\cite{law2018cornernet},
require one inference to end-to-end localize and categorize objects,
which are effective in latency but hard to cover all objects.
Two-stage methods, \eg R-CNN~\cite{girshick2014rich}, Faster-RCNN~\cite{fastrrcnn}, and Cascade R-CNN~\cite{Cai_2019},
decompose object detection
into region proposals generation
and objects detection from the proposal,
enhancing the detection quality at the cost of slow inference speed.
Despite the difference,
object detectors tend to enhance the multi-scale ability
to handle objects of various sizes~\cite{liu2016ssd,fastrrcnn,Cai_2019,chen2019hybrid}.
SSD~\cite{liu2016ssd} merges features from multiple stages to detect objects.
Faster-RCNN~\cite{fastrrcnn} utilizes a feature pyramid network to aggregate features with multiple scales.
Cascade R-CNN~\cite{Cai_2019} and HTC~\cite{chen2019hybrid} perform cascaded
multi-stage feature fusion and refinement.
We show that the proper receptive field settings for these methods
can further improve the detection ability.

\subsubsection{Instance Segmentation}
Instance segmentation aims to assign category labels
to each pixel of instances~\cite{Hu_2021_CVPR,Shen_2021_CVPR,Ding_2021_CVPR},
which is similar to object detection as they both require localizing objects.
Therefore, common instance segmentation methods add the segmentation branch
on object detectors to segment instances from the bounding boxes,
\ie Mask-RCNN~\cite{He_2017} extends Faster-RCNN by adding an object mask predicting branch.
The multi-scale ability in object detectors, \eg Cascade R-CNN~\cite{Cai_2019}
and HTC~\cite{chen2019hybrid}, is naturally inherited to the instance segmentation.
Some works focus on refining the boundary of segmentation masks~\cite{cheng2020boundary,hayder2017boundary,liang2020polytransform,yuan2020segfix,Tang_2021_CVPR},
which still rely on feature extractors with proper receptive fields.
We observe that the receptive field search benefits the performance of instance segmentation.

\subsection{Network Architecture Search}

The genetic algorithm~\cite{mitchell1998introduction} has achieved remarkable
performance on a wide range of applications.
Many genetic-based methods have been recently introduced for the
neural networks architecture search of vision tasks~\cite{real2019regularized,liu2017hierarchical,xie2017genetic,sun2020automatically,lu2019nsga}.
An evolutionary coding scheme is proposed in Genetic CNN~\cite{xie2017genetic}
to encode the network architecture to a binary string.
A hierarchical representation is presented by Liu~\etal~\cite{liu2017hierarchical}
to constrain the search space.
Real~\etal~\cite{real2019regularized} regularize the evolution by an
age property selection operation.
Sun~\etal~\cite{sun2020automatically} introduce a variable-length
encoding method for effective architecture designing.
However, the genetic algorithm requires the training of each candidate,
consuming too much computational cost when faced with a huge search space.

Differentiable architecture search~\cite{liu2019darts,chu2020fair,xu2019pc,wan2020fbnetv2,chen2019progressive,liu2019auto}
saves the training time
by introducing a large network containing subnetworks
with different searching options.
The importance of searched blocks is
determined by gradient backpropagation~\cite{rumelhart1986learning}.
However, these network architecture search methods are designed for finding
a limited number of operations such as convolution, ReLU,
batch normalization, short connection,~\etc.
Thus, these methods cannot be directly used for receptive field search due to
the different searching targets, \eg they cannot handle the huge receptive field combinations
search space.
Fair DARTS~\cite{chu2020fair} solves the problem of performance collapse due to
the unfair advantage in exclusive competition between different operators.
The receptive field searching has no such problem as it contains the same operation.
\cite{xu2019pc} reduces the memory cost of supernet by randomly sampling a proportion of channels for
operation search and bypassing the held out part in a shortcut. 
This approach is unsuitable for receptive field search as the shortcut does not belong to the receptive field search space.
\cite{wan2020fbnetv2} conducts channel number and feature resolution searching with a masking mechanism for feature map reuse.
The masking mechanism for feature map reuse cannot be applied to the receptive field search represented by dilation rate
because different dilation rates do not belong to each other and cannot be chosen with different masks.
\cite{chen2019progressive} narrows the performance gap between models searched from a small dataset
and evaluated on the large dataset.
Theoretically, \cite{chen2019progressive} is orthogonal to receptive field search space.
However, the high efficiency of the proposed local search allows to directly search on the large dataset
instead of on a small dataset.
In this paper,
we propose a global search to handle the huge search space with sparse sampling.
The expectation-guided iterative local search then transfers the sparse search space of
receptive fields into the dense one for fine-level searching.

This differentiable search idea is further extended \cite{zela2020understanding}
to deal with semantic segmentation \cite{liu2019auto,zhang2021dcnas}, and
other tasks beyond image classification~\cite{cai2018proxylessnas}.
Auto-deeplab~\cite{liu2019auto} and DCNAS~\cite{zhang2021dcnas} focus on searching the feature resolution for different stages
of the semantic segmentation network.
We show that our proposed receptive field searching scheme can find
better receptive fields on these searched segmentation networks.

\begin{figure*}[t!]
	\centering
	\begin{overpic}[width=\linewidth]{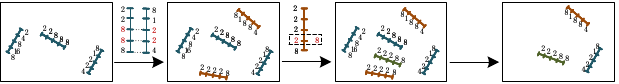}s
		\put(18.6, 1.2){Crossover}
		\put(46, 1.2){Mutation}
		\put(73, 1.2){Selection}
		\put(74, 6){$E(C_i)$}
	\end{overpic}
	\caption{Illustration of one iteration in our genetic-based global search algorithm.
		Step1. Gradually sparse random sampling initial receptive field combinations;
		Step2. Crossover  between  segments  of  the  receptive  field  combinations;
		Step3. Randomly mutating the receptive fields to generate new individuals;
		Step4. Selecting individuals for the next iteration based on estimated performance of models trained with early stopping strategy.
	}
	\label{fig:population}
\end{figure*}

\section{Method}

The pipeline of our proposed global-to-local search method has two components:
(i) a genetic-based global search algorithm that produces coarse but
competitive combinations of the receptive fields;
(ii) an expectation-guided iterative local search scheme that locally refines the global-searched coarse structures.

\subsection{Description} \label{sec:de}

Our objective is to efficiently search for optimal receptive field combinations
for the given dataset.
The receptive field can be represented in multiple forms:
the dilation rate, kernel size, pooling size, stride,
and the stack number of layers.
Our method is initially designed for temporal action segmentation.
We mainly follow the MS-TCN~\cite{farha2019ms} to formulate
the receptive fields using the combinations of dilation
rates in layers and propose to evolve these combinations during the searching process.
Other receptive field representations can also be applied to
the proposed global-to-local search with minor adjustments.
Though we conduct major experiments on the temporal action segmentation task,
our receptive field searching method
can easily be generalized to new tasks, as introduced in~\secref{sec:newtask}.

Suppose a TCN has $L$ convolutional layers and $D=\{d_1,d_2,...,d_N\}$
is the possible dilation-rates/receptive-fields in each layer.
The combination of receptive fields is represented with
$C=\{c_1,...,c_l,...,c_L\}$,
where $l\in [1, L]$ is the index of layers with dilated convolutions,
and $c_l\in D$ is the receptive field of each layer.
There are $|D|^L$ possible combinations of receptive fields,
\ie the possible receptive field combinations in MS-TCN~\cite{farha2019ms} is $1024^{40}$ when dilation rates range from 1 to 1024.
Directly searching for effective combinations in such a large search space is impractical.
We thus decompose the searching process into the global and local search to
find the combination in a coarse-to-fine manner.
\subsection{Global Search} \label{sec:gs}

The global search aims to find possible coarse receptive field combinations in the huge search space that
are very different from human-designed structures,
focusing more on discovering new structures with large diversity to human-designings instead of performance.
To guarantee the diversity of new structures,
we utilize the random sparse sampling strategy and apply a generic algorithm with the random crossover and random mutation
specifically designed for receptive fields.

\myPara{Population initialization with gradually sparse sampling.}
The objective of the global search is to find the coarse receptive field combinations
at an affordable cost.
Therefore, we reduce the search space by sparsely sampling the dilation rates within layers.
Multiple sparse discrete sampling strategies such as uniform sampling,
gradually sparse sampling, and gradually dense sampling can be applied to sparse the search space.
Because small receptive fields benefit the extraction of precise local details,
and large receptive fields contribute to coarse long-term dependencies.
A gradually sparse sampling scheme from small to large dilation rates
is appropriate for common tasks, \eg temporal action segmentation.
Therefore, we formulate the receptive field space in global search as:
\begin{equation}
	D_g=\{d_i = k^{i}, i \in [0,1,\cdots T] \},
	\label{eq:d_g_space}
\end{equation}
where $k$ is the controller of the search space sparsity,
and $T$ determines the largest receptive field.
With the same maximum receptive field, $|D_g| \ll |D|$,
the search space is greatly reduced.
\ie when set $k=2$, and set the maximum receptive field to $1024$ as in MS-TCN,
the search space is reduced from $1024^{40}$ to $11^{40}$.
The population of receptive field combinations can be described as
a group of candidate structures $P=\{C_{i}, i \in [1, M]\}$,
where $C_{i}$ is the candidate structure in the global search space,
and $M$ is the number of individuals in the population.

However, the reduced space of receptive field combinations can still be huge,
and unaffordable for a brute force search.
We propose a genetic algorithm~\cite{mitchell1998introduction} based method
to find coarse combinations that are competitive or even better
than human designing.
We now detail the selection, crossover, and mutation process within
our proposed global search method.

\myPara{Selection according to early stopped training.}
We need to select samples from
the population of receptive field combinations $P$ for each iteration.
The selection operation selects individuals to be kept in $P$ based on the
estimated performance of each structure $C_{i}$, denoted by $E(C_{i})$:
\begin{equation}\label{eq:e_c}
	E(C_{i}) = f(V|C_{i}, \theta_n),
\end{equation}
where $f(\cdot)$ is the task-specific evaluation metrics on the validation set $V$,
\eg frame-wise accuracy for temporal action segmentation, $\theta_n$ is a model trained with $n$ epochs.
The major cost of the global search is the performance evaluation of candidate structures.
The global search aims to find coarse structures and reasonable performance,
allowing searched structures to have sub-optimal performance.
Also, we observe the receptive field combinations play a key role in the model convergence,
\ie a model with good receptive fields converges much faster than a model equipped with bad receptive fields.
To reduce evaluation cost,
we choose to early stop the training of candidate structures when the trained models
can roughly show the relative performance gap of different structures,
\eg training MS-TCN for 5 epochs can reflect the structure performance.
The early stop training strategy substantially reduces the structure evaluation cost.

\myPara{Crossover between segments of the receptive field combination.}
This operation generates new samples of receptive field combinations.
Every two combinations in the population are exchanged to bear new patterns of
the combination while maintaining the local structures.
Each $C_{i}$ will be selected for the crossover operation with probability $p(C_{i})$:
\begin{equation}\label{eq:p_c}
	p(C_{i})= \frac{E(C_{i})}{\sum_{i}^{M}E(C_{i})}.
\end{equation}
Since the representation ability lies in the combination patterns,
we want to reserve the local combination patterns during the crossover.
Instead of randomly exchanging individual points,
we choose to exchange random segments of the receptive field combination.
Specifically, we randomly choose two anchors and exchange receptive field combination segments within two anchors
to generate new samples.

\myPara{Random receptive field mutation.}
The mutation operation avoids getting stuck
in local optimal results by choosing an
individual with pre-defined probability $p_m \in [0,1]$ and
randomly changes each value within the selected combination with
pre-defined probability $p_{s} \in [0,1]$.
To reduce the searching cost, we also apply the gradually sparse sampling strategy
when choosing a new receptive field value.

\begin{algorithm}[htb]
	\caption{Global Search.}
	\label{alg:global_search}
	\begin{algorithmic}
		\renewcommand{\algorithmicrequire}{\tbf{Input:}}
		\Require Iterations $N$, training epoch $n$,
		mutation probability $p_m$, and population size $M$;
		\State Gradually sparse random sampling initial receptive field combinations $P$;
		\For{iter in $[1, N]$}
		\State Selecting individuals for the crossover with the probability obtained with estimated performance in \eqref{eq:p_c};
		\State Crossover between random segments of every two selected receptive field combinations;
		\State Randomly choose combinations with probability $p_m$, and mutate the receptive fields with probability $p_s$
		within sparse sampling search space to generate new individuals;
		\State Training each individual with early stopped $n$ epochs to save evaluation cost;
		\State Selecting the top $M$ individuals based on estimated performance in \eqref{eq:e_c}
		as the new population $P$;
		\EndFor \\
		\Return $P$.
	\end{algorithmic}
\end{algorithm}

The global search process can be summarised as \algref{alg:global_search},
and the illustration of one iteration in global search is given in~\figref{fig:population}.
With the coarse search space and the global search method,
we can find receptive field combinations with different patterns than
human-designed structures while having similar or even better performance.
We further propose the local search to locally find the more efficient
combinations on top of the global-searched structures.
We show in \tabref{tab:ablation_local_init} that local search heavily relies
on the initial structure,
revealing the importance of global search.


\subsection{Expectation-Guided Iterative Local Search} \label{sec:ls}

The local search aims to find more efficient receptive field combinations
at a fine-grained level at a low cost.
A naive approach is to sample finer-grained dilation rates near the
initial dilation rate searched by the global search and
apply existing DARTS algorithms~\cite{cai2018proxylessnas,liu2019darts}
to choose the proper one.
However, even with the good initial structure provided by the global search,
the available range of fine-grained dilation rates is still large.
Existing search algorithms are designed to search sparse operators
with several choices in each layer,
thus can not handle dilation rates with hundreds of choices.
While too sparsely sampling conflicts with our goal of searching
for the finer-grained receptive fields.
Also, DARTS methods search operators with different functionality~\cite{liu2019darts},
while the searching on receptive fields only contains one functional dimension.
Different subsets in the dataset sometimes prefer different searching options.
Searching within a functional dimension enables us to
determine dilation rates with the expectation of all subsets
instead of choosing the option required by one majority subset.
Therefore, we propose an expectation-guided iterative (EGI)
local search scheme to determine the finer-level dilation rates
on top of the global-searched structures.

Suppose that the receptive field of a layer $l$ is $D_{l}$.
For a dataset, once we get the probability mass distribution of dilation rates around $D_{l}$,
we can obtain the expected dilation rate with the weighted average
of the dilation rates required by all subsets.
However, the probability mass of dilation rates for the dataset is inaccessible.
Therefore, we utilize a convolutional weight-sharing scheme to enforce
the learned importance coefficients of dilation rates to approximate the probability mass.
To get the approximated probability mass function of dilation rates,
we first evenly sample $S$ dilation rates near the initial dilation rate
$D_{l}$ within the range of $[D_l \pm \Delta D_{l}]$.
The set of available dilation rates within this layer is
$T_l = \{ d_i | i \in [1, S] \}$,
where $d_i = D_{l}-\Delta D_{l} + (i-1) \cdot 2 \Delta D_{l}/(S-1)$.
$\Delta D_{l}$ is the finer controller of the search space that results in a more dense
sampling than that of the global search.

\begin{figure}[t]
	\centering
	\begin{overpic}[width=\linewidth]{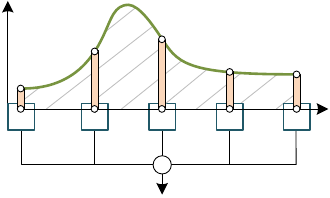}
		\put(4, 22.5){$d_{0}$}
		\put(27, 22.5){$d_{1}$}
		\put(48, 22.5){$d_{2}$}
		\put(68, 22.5){$d_{3}$}
		\put(88, 22.5){$d_{S}$}
		\put(7, 16){$\alpha_{0}$}
		\put(29, 16){$\alpha_{1}$}
		\put(50, 16){$\alpha_{2}$}
		\put(71, 16){$\alpha_{3}$}
		\put(91, 16){$\alpha_{S}$}
		\put(47.9, 9.1){\tbf{+}}
	\end{overpic}
	\caption{The approximated probability mass function of dilation rates is determined by the multi-dilated
		convolutional layer with shared convolutional weights. $d_i$ is the dilation rate, and $\alpha_i$ is the PMF in~\eqref{equ:normalize}.
	}
	\label{fig:local_conv}
\end{figure}

\begin{algorithm}[htb]
	\caption{Expectation-Guided Iterative Local Search.}
	\label{alg:local_search}
	\renewcommand{\algorithmicrequire}{\tbf{Input:}}
	\begin{algorithmic}
		\Require Iterations $N$, initial receptive fields $D$;
		\State Initialize model using given $D$;
		\For{iter in $[1, N]$}
		\State Construct $T_l$ for each layer based on $D$ and initialize $W$ with the same value;
		\State Train model to get the $PMF$ in ~\eqref{equ:normalize};
		\State Obtain new dilation rates through \eqref{equ:recombined};
		\State Update $D$;
		\EndFor \\
		\Return local-searched $D$.
	\end{algorithmic}
\end{algorithm}

\begin{figure}[t!]
	\centering
	\begin{overpic}[width=\linewidth]{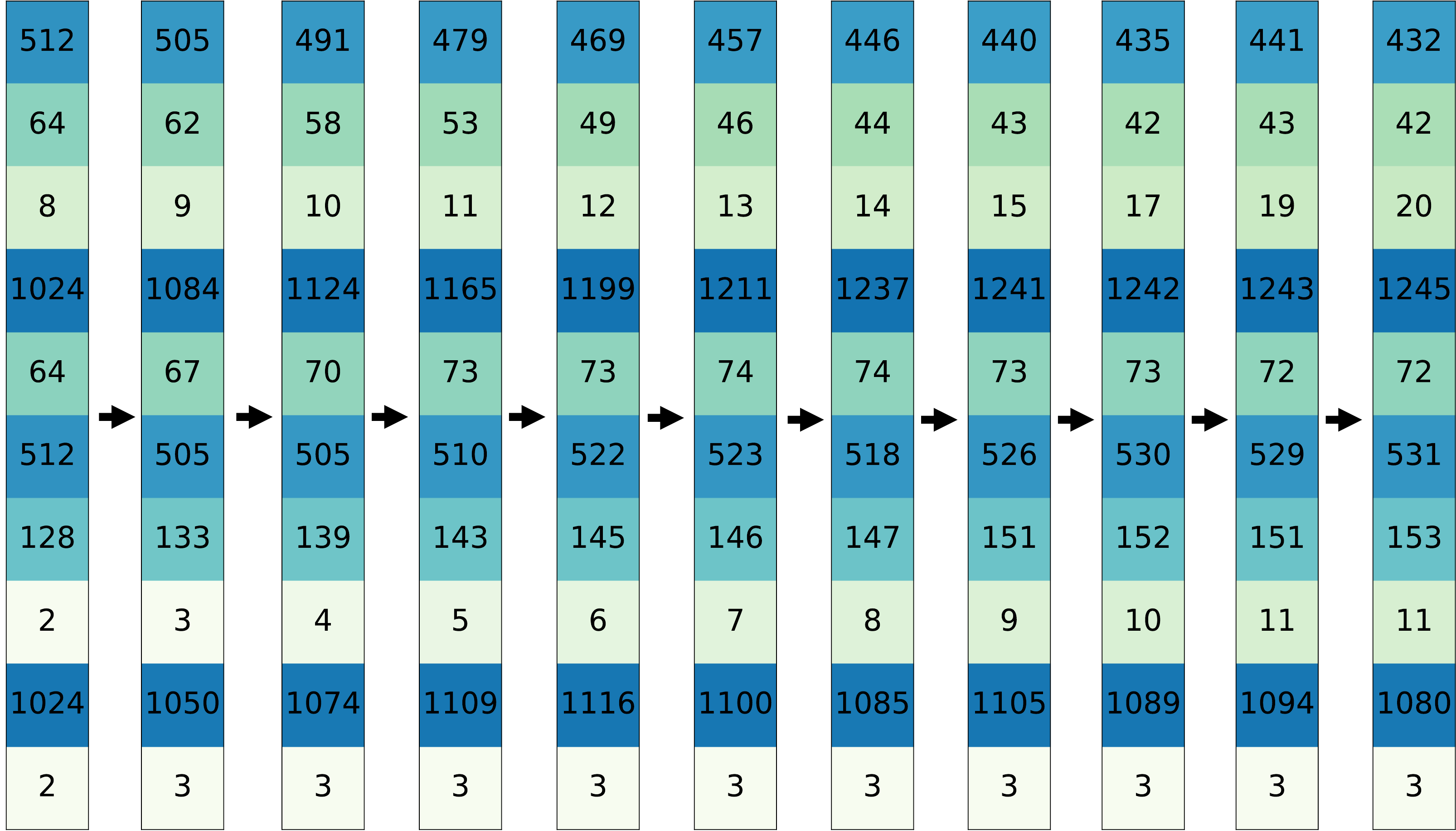}
		\put(46.5, 59){Output}
		\put(47, -4.5){Input}
	\end{overpic}
	\caption{Visualization of receptive field combinations changes during the EGI  local searching process.}
	\label{fig:dilation_changes}
\end{figure}

\def\fMeaures{\multicolumn{3}{c}{F@\{10,25,50\}}}

\begin{table*}[t]
	\centering
	\caption{Performance of the global and local searching stages of
		our global-to-local
		searching method using MS-TCN~\cite{farha2019ms} as the baseline.
		The global search finds new receptive field combinations that are better than baseline.
		Local search further refines the global searched structures to achieve better performance.
	}
	\setlength{\tabcolsep}{1.85mm}
	\begin{tabular}{lccccc|ccccc|ccccc}
		\toprule                            & \multicolumn{5}{c|}{\tbf{BreakFast}} &
		\multicolumn{5}{c|}{\tbf{50Salads}} &
		\multicolumn{5}{c}{\tbf{GTEA}}
		\\
		                                    & F@0.1                                & F@0.25     & F@0.5      & Edit       & Acc        & F@0.1 & F@0.25 & F@0.5 & Edit & Acc & F@0.1 & F@0.25 & F@0.5 & Edit & Acc
		\\ \Xhline{0.8pt}
		{MS-TCN~\cite{farha2019ms}}
		                                    & 52.6                                 & 48.1       & 37.9       & 61.7       & 66.3
		                                    & 76.3                                 & 74.0       & 64.5       & 67.9       & 80.7
		                                    & 87.5                                 & 85.4       & 74.6       & 81.4       & 79.2
		\\
		{Reproduce}
		                                    & 69.1                                 & 63.7       & 50.1       & 69.9       & 67.3
		                                    & 78.8                                 & 75.3       & 64.4       & 71.4       & 77.8
		                                    & 87.1                                 & 83.6       & 70.4       & 81.1       & 75.5
		\\ \hline
		{Global}
		                                    & 72.2                                 & 66.0       & 51.5       & 71.0       & 69.2
		                                    & 79.3                                 & 76.5       & 68.1       & 71.9       & 81.2
		                                    & 89.1                                 & 87.1       & 74.4       & 84.2       & \tbf{78.6}
		\\
		{Global+Local}
		                                    & \tbf{74.9}                           & \tbf{69.0} & \tbf{55.2} & \tbf{73.3} & \tbf{70.7}
		                                    & \tbf{80.3}                           & \tbf{78.0} & \tbf{69.8} & \tbf{73.4} & \tbf{82.2}
		                                    & \tbf{89.9}                           & \tbf{87.3} & \tbf{75.8} & \tbf{84.6} & 78.5                                                                                   \\
		\bottomrule
	\end{tabular}
	\label{tab:search}
\end{table*}

With the dilation rates set $T_l$,
we propose a multi-dilated layer composed of a shared
convolutional weight and multiple branches with different dilation rates,
as shown in~\figref{fig:local_conv}.
Each branch has a unique coefficient to determine the importance of the
dilation rate.
During the searching process,
the coefficients are updated with the gradient backpropagation to reflect the receptive field requirements of the dataset.
Existing DARTS schemes~\cite{liu2019darts,zela2020understanding} have separated operator weights in each branch.
In contrast, our convolutional weight-sharing strategy forces the model to learn the
approximated probability of receptive fields and ease the model convergence.
Specifically, the dilation rates in the multi-dilated convolutional layer
are set to $T_l$.
Apart from the shared convolutional $\theta$,
the multi-dilated layer contains coefficient
$W = \{w_1, w_2, ..., w_i, i \in [1, S]\}$
to determine the importance of the dilation rates.
Both $\theta$ and $W$ are learnable parameters and can be trained with gradient backpropagation.
For each iteration, each value in $W$ is reinitialized with the same initial value.

$W$ is unbounded,
thus cannot be directly used to determine the dilation rates probability.
Therefore, we propose a normalization function to get the approximated probability mass
function $PMF(d_i)$ of dilation rates through normalizing $w_i$:
\begin{equation}\label{equ:normalize}
	PMF(d_i) = \alpha_i = \frac{|w_i|}{\sum_{i}^{S} |w_i| }.
\end{equation}
With the probability mass function, given the input feature $x$,
the output $y$ of the multi-dilated convolutional layer can be written
as follows:
\begin{equation}\label{equ:forward}
	y = \sum_{i}^{S} \alpha_i \Psi(x, d_i, \theta),
\end{equation}
where $\Psi(x, d_i, \theta)$ is the convolutional operation
with the shared convolutional weight $\theta$ and dilation rate $d_i$.
$\alpha_i$ is updated with gradient optimization.
Once we get the probability mass function,
the newly searched dilation rate $D_l^{'}$ is obtained with the expectation:
\begin{equation}\label{equ:recombined}
	D_l^{'} = \lfloor \sum_{d_{i}\in T_l} PMF(d_i) \cdot d_{i} \rfloor.
\end{equation}
To reduce the computational cost during the local search process,
we reduce the number of dilation rates in $T_l$ to 3 by default
and apply the iterative search scheme to find the more suitable dilation rate
based on the  $D_l^{'}$ from the last iteration.
The local search process can be summarised as \algref{alg:local_search}.
Furthermore,~\figref{fig:dilation_changes} visualizes the dilation rates
changes during the local searching process.

\myPara{Parallel receptive fields for multi-scale enhancement.}
The local search results in one dilation rate for each convolution.
However, we observe that some spatial tasks, \eg instance segmentation and object detection,
require the parallel multi-scale ability to process small and large objects in the senses.
Our expectation-guided local search scheme can provide the parallel multi-scale ability
with different dilation rates and shared convolutional weights.
Therefore, we extend the local-searched structure to the parallel version
by keeping the dilation rates in $T_l$ instead of merging them after the last iteration of searching.
The parallel version only has $|T_l|$ extra parameters compared with the single branch version.
The parallel structures have significant improvement over the single branch structures
in instance segmentation and object detection tasks.

\def\GTEA{{GTEA}~\cite{fathi2011learning}}
\def\Salads{{50Salads}~\cite{stein2013combining}}
\def\BreakFast{{BreakFast}~\cite{kuehne2014language}}
\begin{table}[tb]
	\centering
	\caption{Details of three temporal action segmentation datasets.
		\#Cls and \#Vid are the numbers of classes and videos, respectively.
		\#Frame is the average frame of videos.
	}
	\setlength{\tabcolsep}{3.1mm}
	\begin{tabular}{lcccc}  \toprule
		           & \#Cls & \#Vid & \#Frame & Scene             \\ \midrule
		\GTEA      & 11    & 28    & 1115    & daily activities  \\ \hline
		\Salads    & 17    & 50    & 11552   & preparing salads  \\   \hline
		\BreakFast & 48    & 1712  & 2097    & cooking breakfast \\ \bottomrule
	\end{tabular}
	\label{tab:datasets}
\end{table}
\subsection{RF-Next: Next Generation Receptive Field Models}
\label{sec:newtask}
Our global-to-local receptive field searching scheme is
suitable for various models that utilize convolutions.
Given an initial network structure,
we apply the searching scheme to convolutions with kernel sizes larger than one.
For easy implementation, we utilize dilation rates
to represent the receptive field.
The global search aims to find receptive field combinations
beyond human knowledge, which is optional as
many models have been manually tuned.
The local search finds suitable fine-grained receptive fields
with a small extra cost, and thus it can be easily applied
to human-designed models of various tasks.
Enhanced by our receptive field search,
these \textbf{Next} generation \textbf{R}eceptive \textbf{F}ield models, namely RF-Next models,
show advantages on many tasks, \eg object detection,
instance segmentation, semantic segmentation,
speech synthesis, and sequence modeling.

\section{Experiments on Temporal Action Segmentation} \label{sec:experimets}
Temporal action segmentation requires a relatively large range of receptive fields,
which is suitable for verifying the effectiveness of our proposed global-to-local search.
Therefore, we do our major experiments on the temporal action segmentation task.
This section introduces the implementation details of our proposed global-to-local search scheme
and shows the superiority of searched receptive field combinations over the human-designed patterns
on the temporal action segmentation task.
We also give an analysis of the search scheme and the property of searched structures.
\subsection{Implementation Details}
\label{sec:dataset}
\myPara{Structure Searching and Training.}
Our proposed method is implemented with the PyTorch~\cite{paszke2019pytorch}, and Jittor~\cite{hu2020jittor} frameworks.
Following existing works~\cite{MS-TCN-PAMI20,farha2019ms}, features are first extracted from videos using the I3D network~\cite{carreira2017quo} and then
passed to temporal action segmentation models to get the temporal
segmentation.
Since our proposed global-to-local search scheme is model-agnostic,
the training settings for model evaluation,
\ie training epochs, optimizer, learning rate, batch size, keep the same with
the cooperation methods~\cite{MS-TCN-PAMI20,wangboundary,chen2020action}. %
In the global search stage, we set the total iterations $N = 100$,
$k=2$ in~\eqref{eq:d_g_space}, the initialized population size $M=50$,
and mutation probability $p_m$=$p_s$=$0.2$.
The $T$ in \eqref{eq:d_g_space} is set to 10, indicating the maximum dilation rate of the global search space is 1024.
We observe that 5 epochs of training can reflect the structure performance,
and therefore models are trained with 5 epochs for evaluation. In the EGI local search stage,
$\Delta D_l$ and $S$ are set to be $0.1D_l$ and 3, respectively. We train the model for 30 epochs during
local search, and each iteration contains 3 epochs.

\myPara{Datasets.}
Following~\cite{farha2019ms,MS-TCN-PAMI20,wangboundary,chen2020action},
we evaluate our proposed method on three popular temporal action segmentation datasets:
Breakfast~\cite{kuehne2014language}, 50Salads~\cite{stein2013combining}, and GTEA~\cite{fathi2011learning}.
The details of the three datasets are summarized in~\tabref{tab:datasets}.
As far as we know, the Breakfast dataset is the
largest public dataset for the temporal action segmentation task, which has a larger number of categories and samples
compared with the other two datasets.
So we perform our ablations mainly on the Breakfast dataset if not otherwise stated.
Following common settings~\cite{farha2019ms,MS-TCN-PAMI20,wangboundary,chen2020action},
we perform 4-fold cross-validation
for the Breakfast and GTEA datasets and 5-fold cross-validation for the 50Salads dataset.

\myPara{Evaluation Metrics.}
We follow previous works~\cite{farha2019ms,MS-TCN-PAMI20,wangboundary,chen2020action}
to use the frame-wise accuracy (Acc), segmental edit score (Edit)~\cite{lea2017temporal},
and segmental F1 score~\cite{li2015convolutional} at the temporal intersection over union with thresholds 0.1, 0.25, 0.5 (F@0.1, F@0.25, F@0.5)
as our evaluation metrics.

\begin{figure*}[!t]
	\centering
	\small
	\begin{overpic}[width=\linewidth]{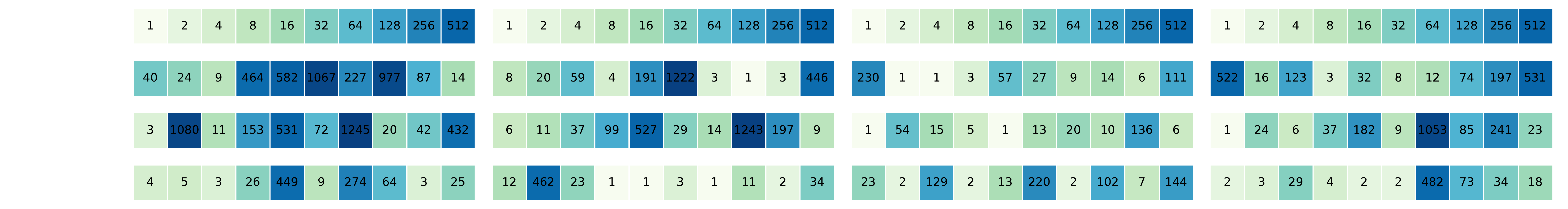}
		\put(0,11.1){Baseline}
		\put(0,7.9){50Salads}
		\put(0,4.5){BreakFast}
		\put(0,1){GTEA}
	\end{overpic}
	\caption{Visualization of the global-to-local searched structures of three datasets with the MS-TCN baseline.
		Each row represents the dilations of one structure, which contains four stages.}
	\label{fig:architectures_datasets}
\end{figure*}


\begin{table}[t]
	\centering
	\caption{Cooperation with existing temporal action segmentation methods.
		We perform the whole search pipeline based on MS-TCN~\cite{farha2019ms}.
		Because of the limited computing resources,
		we only perform the EGI local search on MS-TCN++~\cite{MS-TCN-PAMI20} and
		BCN~\cite{wangboundary}, denoted by \dag. SSTDA~\cite{chen2020action} uses MS-TCN~\cite{farha2019ms} as a backbone,
		so we directly add our searched
		structure to SSTDA, denoted by \ddag.}
	\setlength{\tabcolsep}{2.6mm}
	\begin{tabular}{lccccc}  \toprule
		\tbf{BreakFast}
		 & F@0.1 & F@0.25 & F@0.5 & Edit & Acc  \\
		\midrule
		{ED-TCN~\cite{lea2017temporal}}
		 & -     & -      & -     & -    & 43.3 \\
		{HTK (64)~\cite{kuehne2016end}}
		 & -     & -      & -     & -    & 52.0 \\
		{TCFPN~\cite{ding2018weakly}}
		 & -     & -      & -     & -    & 56.3 \\
		{GRU~\cite{richard2017weakly}}
		 & -     & -      & -     & -    & 60.6 \\
		{GTRM~\cite{huang2020improving}}
		 & 57.5  & 54.0   & 43.3  & 58.7 & 65.0 \\
		\hline
		{MS-TCN~\cite{farha2019ms}}
		 & 52.6  & 48.1   & 37.9  & 61.7 & 66.3 \\
		{RF-MS-TCN}
		 & 74.9  & 69.0   & 55.2  & 73.3 & 70.7 \\
		\hline
		{MS-TCN++~\cite{MS-TCN-PAMI20}}
		 & 64.1  & 58.6   & 45.9  & 65.6 & 67.6 \\
		{RF\dag-MS-TCN++}
		 & 72.4  & 66.8   & 53.5  & 70.2 & 69.6 \\
		\hline
		{BCN~\cite{wangboundary}}
		 & 68.7  & 65.5   & 55.0  & 66.2 & 70.4 \\
		{RF\dag-BCN}
		 & 72.5  & 69.9   & 60.2  & 69.0 & 72.9 \\
		\hline
		{SSTDA~\cite{chen2020action}}
		 & 75.0  & 69.1   & 55.2  & 73.7 & 70.2 \\
		{RF\ddag-SSTDA}
		 & 76.3  & 69.9   & 54.6  & 74.5 & 70.8 \\
		\bottomrule
	\end{tabular}
	\label{tab:compare_sota}
\end{table}

\begin{table}[t]
	\centering
	\caption{Comparison with existing temporal action segmentation methods on the 50Salads and GTEA datasets.}
	\setlength{\tabcolsep}{2.7mm}
	\begin{tabular}{lccccc}  \toprule
		\tbf{50Salads}
		 & F@0.1      & F@0.25     & F@0.5      & Edit       & Acc        \\
		\midrule
		{Spatial CNN~\cite{lea2016segmental}}
		 & 32.3       & 27.1       & 18.9       & 24.8       & 54.9       \\
		{Bi-LSTM~\cite{singh2016multi}}
		 & 62.6       & 58.3       & 47.0       & 55.6       & 55.7       \\
		{Dilated TCN~\cite{lea2017temporal}}
		 & 52.2       & 47.6       & 37.4       & 43.1       & 59.3       \\
		{ST-CNN~\cite{lea2016segmental}}
		 & 55.9       & 49.6       & 37.1       & 45.9       & 59.4       \\
		{TUnet~\cite{ronneberger2015u}}
		 & 59.3       & 55.6       & 44.8       & 50.6       & 60.6       \\
		{ED-TCN~\cite{lea2017temporal}}
		 & 68.0       & 63.9       & 52.6       & 59.8       & 64.7       \\
		{TResNet~\cite{he2016deep}}
		 & 69.2       & 65.0       & 54.4       & 60.5       & 66.0       \\
		{TricorNet~\cite{ding2017tricornet}}
		 & 70.1       & 67.2       & 56.6       & 62.8       & 67.5       \\
		{TRN~\cite{lei2018temporal}}
		 & 70.2       & 65.4       & 56.3       & 63.7       & 66.9       \\
		{TDRN~\cite{lei2018temporal}}
		 & 72.9       & 68.5       & 57.2       & 66.0       & 68.1       \\
		{MS-TCN++~\cite{MS-TCN-PAMI20}}
		 & 80.7       & 78.5       & 70.1       & 74.3       & 83.7       \\
		\hline
		{MS-TCN~\cite{farha2019ms}}
		 & 76.3       & 74.0       & 64.5       & 67.9       & 80.7       \\
		{RF-MS-TCN}
		 & \tbf{80.3} & \tbf{78.0} & \tbf{69.8} & \tbf{73.4} & \tbf{82.2} \\ \hline
		BCN~\cite{wangboundary}
		 & 82.3       & 81.3       & 74.0       & 74.3       & 84.4       \\
		RF-BCN
		 & \tbf{85.8} & \tbf{83.6} & \tbf{76.5} & \tbf{78.1} & \tbf{85.5} \\
		\bottomrule
		\toprule
		\multirow{-2}{*}{}
		\textbf{GTEA}
		 & F@0.1      & F@0.25     & F@0.5      & Edit       & Acc        \\
		\midrule
		{Spatial CNN~\cite{lea2016segmental}}
		 & 41.8       & 36.0       & 25.1       & -          & 54.1       \\
		{Bi-LSTM~\cite{singh2016multi}}
		 & 66.5       & 59.0       & 43.6       & -          & 55.5       \\
		{Dilated TCN~\cite{lea2017temporal}}
		 & 58.8       & 52.2       & 42.2       & -          & 58.3       \\
		{ST-CNN~\cite{lea2016segmental}}
		 & 58.7       & 54.4       & 41.9       & -          & 60.6       \\
		{TUnet~\cite{ronneberger2015u}}
		 & 67.1       & 63.7       & 51.9       & 60.3       & 59.9       \\
		{ED-TCN~\cite{lea2017temporal}}
		 & 72.2       & 69.3       & 56.0       & -          & 64.0       \\
		{TResNet~\cite{he2016deep}}
		 & 74.1       & 69.9       & 57.6       & 64.4       & 65.8       \\
		{TricorNet~\cite{ding2017tricornet}}
		 & 76.0       & 71.1       & 59.2       & -          & 64.8       \\
		{TRN~\cite{lei2018temporal}}
		 & 77.4       & 71.3       & 59.1       & 72.2       & 67.8       \\
		{TDRN~\cite{lei2018temporal}}
		 & 79.2       & 74.4       & 62.7       & 74.1       & 70.1       \\
		{MS-TCN++~\cite{MS-TCN-PAMI20}}
		 & 88.7       & 87.4       & 73.5       & 83.0       & 78.2       \\
		\hline
		{MS-TCN~\cite{farha2019ms}}
		 & 87.5       & 85.4       & 74.6       & 81.4       & 79.2       \\
		{Reproduce}
		 & 87.1       & 83.6       & 70.4       & 81.1       & 75.5       \\
		{RF-MS-TCN}
		 & 89.9       & 87.3       & 75.8       & 84.6       & 78.5       \\ \hline
		BCN~\cite{wangboundary}
		 & 88.5       & 87.1       & 77.3       & 84.4       & 79.8       \\
		RF-BCN
		 & \tbf{92.1} & \tbf{90.2} & \tbf{79.2} & \tbf{87.2} & \tbf{80.6} \\
		\bottomrule
	\end{tabular}
	\label{tab:compare_sota_gtea}
\end{table}

\subsection{Performance Evaluation}
\myPara{Global2Local Search.}
Our proposed global-to-local search aims to find new combinations of receptive fields better than human designings.
We mainly take MS-TCN~\cite{farha2019ms} as our baseline architecture to perform the global-to-local search.
When testing the MS-TCN on the Breakfast dataset,
we train all models with batch size 8 to save training time.
The reproduced results shown in~\tabref{tab:search}
indicate that a large batch size achieves much better performance.
\tabref{tab:search} shows that global-to-local searched structures achieve considerable performance improvements
than human-designed baselines, \ie the searched structure surpasses the reproduced baseline with 5.8$\%$ in terms of F@0.1.
The global-to-local search focuses on receptive field combinations,
thus cooperating with existing SOTA temporal action segmentation methods to improve their performance.
As shown in \tabref{tab:compare_sota}, on the large-scale BreakFast dataset,
global-to-local search consistently boosts the performance
of MS-TCN++~\cite{MS-TCN-PAMI20}, BCN~\cite{wangboundary}, and SSTDA~\cite{chen2020action}.
Also, we give comparisons on two small-scale datasets, 50Salads and GTEA datasets in \tabref{tab:compare_sota_gtea}, proving the effectiveness of our proposed global-to-local search.

\begin{figure}[!t]
	\centering
	\begin{overpic}[width=\linewidth]{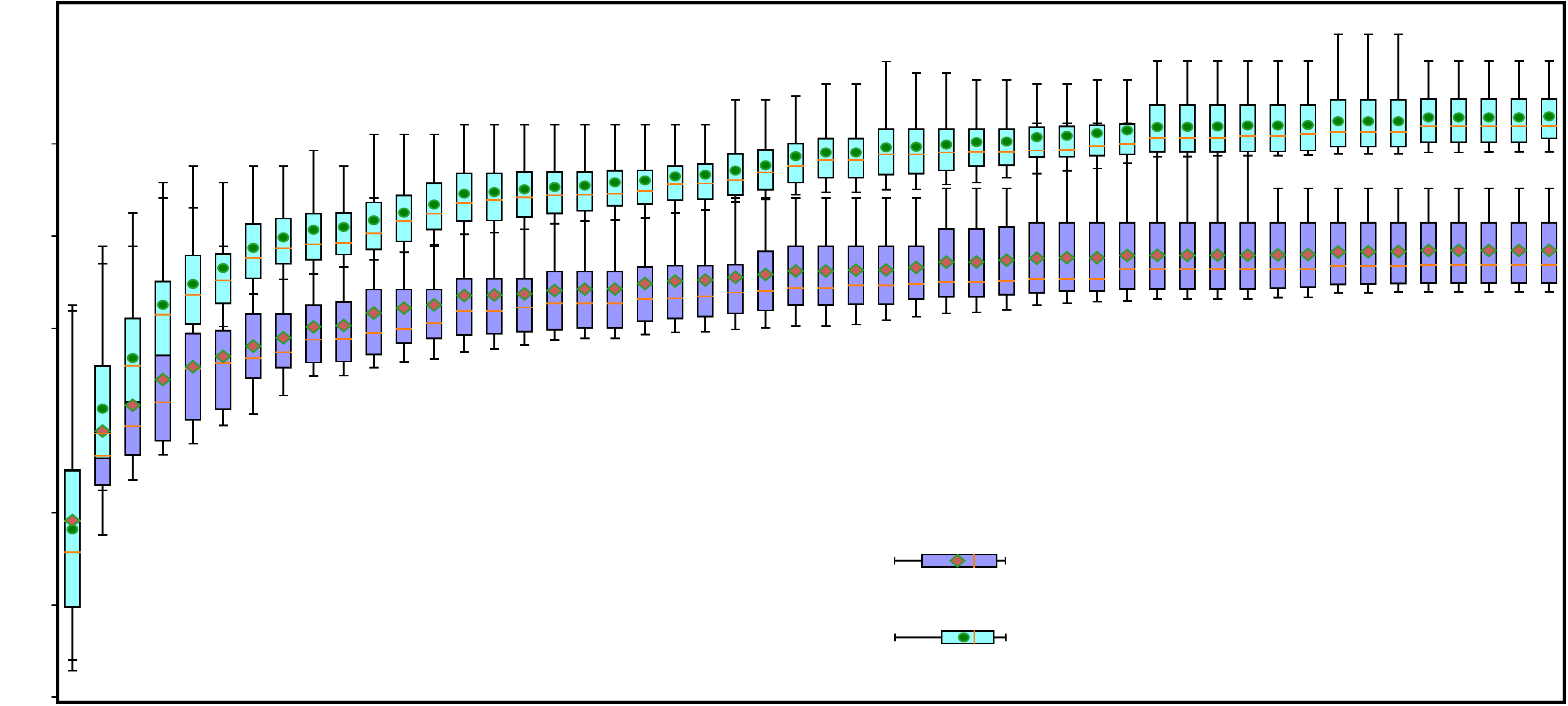}
		\put(66,3){Global Search}
		\put(66,8.5){Random Search}
		\put(-1,16){\rotatebox{90}{F@0.1}}
		\put(45, -3.5){Iterations}
	\end{overpic}
	\caption{Performance comparison between our proposed genetic-based search and random search during the global search stage. }
	\label{fig:random_vs_pop}
\end{figure}

\myPara{Global Search.}
Global search reduces the computational cost with the sparse search space
and our proposed genetic-based searching scheme.
\figref{fig:random_vs_pop} shows the performance change of models during the global searching process.
Compared with the random search, the genetic-based global search convergences faster.
The standard division of model performance searched by genetic-based search is smaller
than the random search, showing the stability of our proposed search scheme.
The visualized well-performed global-searched structures in~\figref{fig:architectures_datasets}
prove that the global search discovers various structures completely different from
human-designed patterns.
\tabref{tab:ablation_local_init} also shows that the local search heavily relies on global-searched structures
to achieve better performance.

\begin{table}
	\caption{Ablation about the proposed EGI local search.}
	\centering
	\subtable[Performance of EGI local search and DARTS-related methods.]{
		\centering
		\setlength{\tabcolsep}{2.2mm}
		\begin{tabular}{lccccc}  \toprule
			\tbf{BreakFast}
			 & F@0.1      & F@0.25     & F@0.5      & Edit       & Acc        \\
			\midrule
			DARTS~\cite{liu2019darts}
			 & 73.4       & 67.3       & 53.1       & 72.7       & 69.5       \\
			+Early Stop~\cite{liang2019darts+}
			 & 73.8       & 67.6       & 52.8       & 72.8       & 69.3       \\
			\midrule
			DARTS+Early Stop
			 & 73.8       & 67.6       & 52.8       & 72.8       & 69.3       \\
			+ Fair DARTS~\cite{chu2019fairdarts}
			 & 73.3       & 67.5       & 52.9       & 71.9       & 69.9       \\
			\midrule
			{Ours}
			 & \tbf{74.9} & \tbf{69.0} & \tbf{55.2} & \tbf{73.3} & \tbf{70.7} \\
			\bottomrule
		\end{tabular}
		\label{tab:ablation_das}
	}
	\\
	\subtable[Performance of EGI local search initialized by different structures.]{
		\centering
		\setlength{\tabcolsep}{2.7mm}
		\begin{tabular}{lccccc}  \toprule
			\tbf{BreakFast}
			 & F@0.1 & F@0.25 & F@0.5 & Edit & Acc  \\
			\midrule
			{random}
			 & 67.7  & 61.8   & 48.3  & 68.4 & 67.0 \\
			{random + local}
			 & 73.6  & 67.8   & 53.7  & 72.3 & 69.9 \\
			\hline
			{baseline~\cite{farha2019ms}}
			 & 69.1  & 63.7   & 50.1  & 71.0 & 69.2 \\
			{baseline + local}
			 & 74.1  & 68.5   & 55.3  & 72.3 & 70.2 \\
			\hline
			{global}
			 & 72.2  & 66.0   & 51.8  & 71.5 & 69.4 \\
			{global + local}
			 & 74.9  & 69.0   & 55.2  & 73.3 & 70.7 \\
			\bottomrule
		\end{tabular}
		\label{tab:ablation_local_init}
	}
	\subtable[Performance of EGI local search initialized by different structures.]{
		\centering
		\setlength{\tabcolsep}{3.2mm}
		\begin{tabular}{lccccc}  \toprule
			\tbf{BreakFast}
			 & F@0.1 & F@0.25 & F@0.5 & Edit & Acc  \\
			\midrule
			{$S=2$}
			 & 74.8  & 68.9   & 55.0  & 73.4 & 70.4 \\
			{$S=3$}
			 & 74.9  & 69.0   & 55.2  & 73.3 & 70.7 \\
			{$S=4$}
			 & 74.9  & 68.8   & 55.1  & 73.3 & 70.9 \\
			\bottomrule
		\end{tabular}
		\label{tab:ablation_s}
	}
	\subtable[Ablation of possible probability mass functions in EGI local search.]{
		\centering
		\setlength{\tabcolsep}{3.2mm}
		\begin{tabular}{lccccc}  \toprule
			\tbf{BreakFast}
			 & F@0.1      & F@0.25     & F@0.5      & Edit       & Acc        \\
			\midrule
			{sigmoid}
			 & 72.7       & 66.9       & 52.7       & 71.8       & 69.4       \\
			{softmax}
			 & 73.2       & 67.2       & 52.0       & 71.6       & 69.7       \\
			{\eqref{equ:normalize}}
			 & \tbf{74.9} & \tbf{69.0} & \tbf{55.2} & \tbf{73.3} & \tbf{70.7} \\
			\bottomrule
		\end{tabular}
		\label{tab:ablation_weight_function}
	}
\end{table}

\begin{table}[t]
	\caption{GPU hours of the global and local search on different temporal action segmentation datasets
		using the RTX 2080Ti GPU based on the MS-TCN method.}
	\centering
	\setlength{\tabcolsep}{2.3mm}
	\begin{tabular}{lccccc}  \toprule
		\textbf{GPU Hours} & \textsl{BreakFast} & \textsl{50Salads} & \textsl{GTEA} \\ \midrule
		{Global Search}    & 144h               & 9h                & 1h            \\
		{Local Search}     & 2.2h               & 0.15h             & 0.05h         \\ \midrule
		MS-TCN Training    & 2.0h               & 0.14h             & 0.05h         \\
		\bottomrule
	\end{tabular}
	\vspace{2pt}
	\label{tab:gpu_hours}
\end{table}

\myPara{Local Search.}
Based on the global-searched structures, our proposed EGI local search aims to fine-tune the receptive field in a
finer search space.
We compare the DARTS~\cite{liu2019darts} method and EGI local search based on the global searched structures, as shown in~\tabref{tab:ablation_das}.
Compared with the DARTS method that only supports several search options,
the EGI local search iteratively finds the accurate dilations in a dense space,
obtaining structures with better performance.
We also compare the EGI local search with some variants of DARTS, \ie the early stop scheme~\cite{liang2019darts+} and Fair DARTS~\cite{chu2019fairdarts}, and our method outperform them with a clear margin.
The early stop scheme improves the performance of DARTS that searched with more epochs, while Fair DARTS~\cite{chu2019fairdarts} has no advantage
on the receptive field search task.
The early stop scheme~\cite{liang2019darts+} stops the searching before the
overfitting of DARTS.
Fair DARTS~\cite{chu2020fair} solves the problem of performance collapse caused by the unfair advantage in exclusive competition between different operator paths,
while the dilation search has no such problem because each path has the same operation.

As shown in \tabref{tab:ablation_s},
EGI local search is insensitive to the number of sampling dilation rates $S$,
as it searches dilation rates with the expectation.
\tabref{tab:ablation_local_init} shows that the EGI local search
can boost the performance of randomly generated, human-designed, and global-searched structures.
Still, the performance of the local-searched structures is related to the initial structures,
as local search focuses on searching for receptive fields within a finer local search space.
We visualize the searching process of the iterative local search in~\figref{fig:dilation_changes}.
The dilation rates for each layer gradually converge to a suitable state during the iterative searching process.
\tabref{tab:ablation_weight_function} verifies different ways to get the approximated probability mass function
$PMF(d_i)$ from
coefficient $w$. \eqref{equ:normalize} is superior to the sigmoid and softmax functions
because it maintains the probability distribution while the other two functions change the distribution non-linearly.

\begin{table}[t]
	\centering
	\caption{Cross-validation performance (F@0.1) of searched structures among the fold 1 of different datasets.
		Arch-dataset indicates the structure is searched on which dataset.}
	\setlength{\tabcolsep}{2.3mm}
	\begin{tabular}{lcccc}
		\toprule
		 & MS-TCN & \tbf{Arch-50Salads} & \tbf{Arch-GTEA} & \tbf{Arch-BF} \\
		\midrule
		\tbf{50Salads}
		 & 67.1   & \tbf{75.4}          & 68.8            & 72.6          \\
		\tbf{GTEA}
		 & 83.8   & 82.4                & \tbf{88.9}      & 85.6          \\
		\tbf{BF}
		 & 69.9   & 75.1                & 72.5            & \tbf{76.4}    \\
		\bottomrule
	\end{tabular}
	\label{tab:ablation_cross_val_datasets}
\end{table}
\begin{figure}[t]
	\centering
	\small
	\begin{overpic}[width=\linewidth]{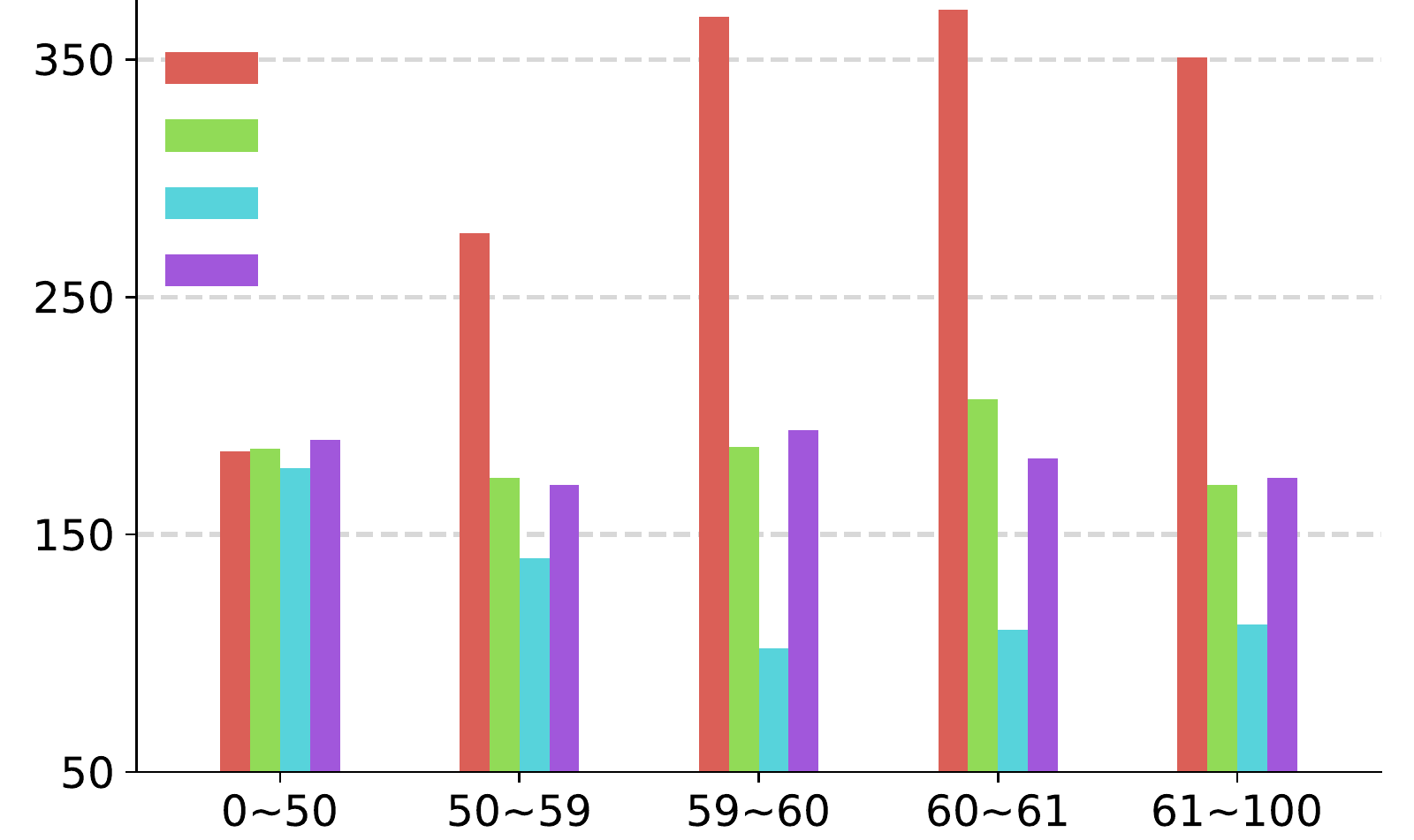}
		\put(20,53.6){stage1}
		\put(20,48.8){stage2}
		\put(20,43.9){stage3}
		\put(20,39.3){stage4}

		\put(32,-4){The range of performance}
		\put(-2,26){\rotatebox{90}{Dilation rate}}
	\end{overpic}
	\caption{Visualization of average dilation rates in each stage and the range of performance of global-searched structures.}
	\label{fig:arverage_dilation}
\end{figure}

\begin{table*}[t]
	\small
	\centering
	\caption{Performance of local search on object detection with COCO~\cite{lin2014microsoft} dataset
		using Faster-RCNN as the baseline method.
		Local-P means the local-searched structure with parallel receptive fields as described in~\secref{sec:ls}.
		-R50 and -R101 denote using ResNet-50 and ResNet-101 as backbones, respectively.
		$S$ indicates using $S$ branches in local search as shown in~\eqref{equ:normalize}.}
	\setlength{\tabcolsep}{1.7mm}
	\begin{tabular}{lccccccc|cccccc}  \toprule
		                                  & \multirow{2}{*}{P} & \multicolumn{6}{c}{validation set} & \multicolumn{6}{c}{test set}                                                                    \\
		                                  &                    & mAP                                & mAP$_{\rm 50}$               & mAP$_{\rm 75}$ & mAP$_{\rm s}$ & mAP$_{\rm m}$ & mAP$_{\rm l}$
		                                  & mAP                & mAP$_{\rm 50}$                     & mAP$_{\rm 75}$               & mAP$_{\rm s}$  & mAP$_{\rm m}$ & mAP$_{\rm l}$                   \\
		\midrule
		Faster-RCNN-R50~\cite{fastrrcnn}  &                    & 37.4                               & 58.3                         & 40.6           & 22.0          & 41.1          & 48.1          &
		37.8                              & 59.0               & 41.0                               & 22.1                         & 40.8           & 46.4                                            \\ \midrule
		+RF ($ S=3$)                      &                    & 39.3                               & 60.6                         & 42.9           & 23.5          & 43.0          & 51.0          &
		39.2                              & 60.9               & 42.6                               & 22.6                         & 41.8           & 48.9                                            \\
		+RF ($ S=3$)                      & \checkmark         & 40.2                               & 61.7                         & 43.8           & 23.5          & 43.9          & 52.2          &
		40.4                              & 62.1               & 44.0                               & 23.6                         & 43.0           & 50.5                                            \\
		+RF ($S=2$)                       &                    & 39.1                               & 60.5                         & 42.3           & 23.2          & 42.8          & 50.3          &
		39.1                              & 60.8               & 42.3                               & 22.7                         & 41.7           & 48.7                                            \\
		+RF ($ S=2$)                      & \checkmark         & 40.0                               & 61.4                         & 43.8           & 23.9          & 43.6          & 51.9          &
		40.3                              & 62.1               & 43.9                               & 23.7                         & 43.0           & 50.4                                            \\
		\midrule
		Faster-RCNN-R101~\cite{fastrrcnn} &                    & 39.4                               & 60.1                         & 43.1           & 22.4          & 43.7          & 51.1          &
		39.7                              & 60.7               & 43.2                               & 22.5                         & 42.9           & 49.9                                            \\
		+RF ($S=3$)                       &                    & 41.1                               & 62.4                         & 44.7           & 24.5          & 45.1          & 53.9          &
		41.2                              & 62.8               & 44.9                               & 23.6                         & 44.1           & 52.0                                            \\
		+RF ($ S=3$)                      & \checkmark         & 42.0                               & 63.2                         & 45.7           & 25.0          & 45.9          & 55.4          &
		42.1                              & 63.8               & 45.8                               & 24.3                         & 45.1           & 53.3                                            \\
		\bottomrule
	\end{tabular}
	\label{tab:object_detection}
\end{table*}
\begin{table}[t]
	\centering
	\caption{Cross-validation performance (F@0.1) of searched structures among different folds of the BreakFast dataset.
		Arch-n means the structure is searched on fold n.}
	\setlength{\tabcolsep}{4.3mm}
	\begin{tabular}{lcccc}  \toprule
		\tbf{BreakFast}
		 & Arch-1 & Arch-2 & Arch-3 & Arch-4 \\
		\midrule
		fold1
		 & 76.4   & 76.3   & 76.2   & 75.7   \\
		fold2
		 & 74.1   & 75.3   & 75.1   & 74.6   \\
		fold3
		 & 76.1   & 76.6   & 76.1   & 75.4   \\
		fold4
		 & 71.7   & 72.1   & 72.0   & 71.8   \\
		\bottomrule
	\end{tabular}
	\label{tab:ablation_cross_val_splits}
\end{table}
\begin{figure*}[!t]
	\centering
	\small
	\begin{overpic}[width=\linewidth]{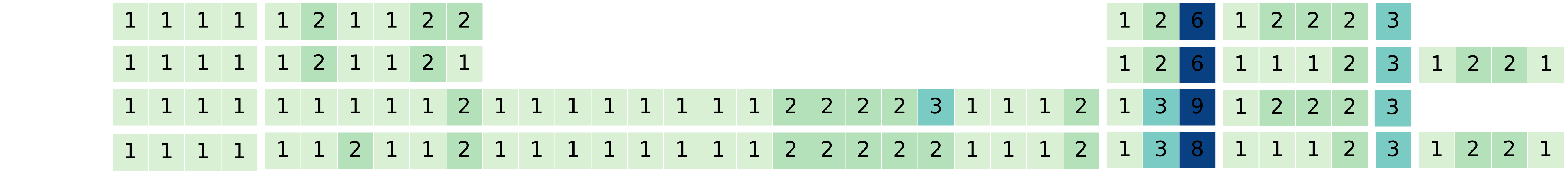}
		\put(0, 9.3){OB-R50}
		\put(0, 6.5){IN-R50}
		\put(0, 3.7){OB-R101}
		\put(0, 0.9){IN-R101}

		\put(11.4, -1.7){S2}
		\put(41.8, -1.7){S3}
		\put(73.3, -1.7){S4}
		\put(81.2, -1.7){FPN}
		\put(87.2, -1.7){RPN}
		\put(93.3, -1.7){MASK}
	\end{overpic}
	\caption{Visualization of the local-searched structures of Faster-RCNN~\cite{fastrrcnn} for object detection (OB) and
		Mask-RCNN~\cite{He_2017} for instance segmentation (IN).
		S2-S4 denotes stage 2 to stage 4 of the ResNet backbone.
		FPN, RPN, and MASK mean the feature pyramid network, region proposal network, and the mask segmentation head
		in Faster-RCNN and Mask-RCNN.}
	\label{fig:architectures_detection}
\end{figure*}

\myPara{Searching Cost.}
We report the cost of our proposed global-to-local search method.
When cooperating with MS-TCN,
the size of the receptive field combination search space is $1024^{40}$.
The cost of searching on such a huge space is unaffordable when using existing search methods.
Our proposed global-to-local search decomposes the searching process into the
global and local search to find the combination in a coarse-to-fine manner.
Since the main bottleneck of the search method is the GPU resources,
we report the GPU hours of the proposed global-to-local search in~\tabref{tab:gpu_hours}.
The global search requires more computational cost to find multiple new well-performed structures
with different patterns than human-designed structures.
The local search needs a small training cost to fine-tune the
global-searched/human-designed structures
in the dense but local search space.

\subsection{Observations}
In this section, we try to exploit the common knowledge contained
in the global-to-local searched structures.

\myPara{Connections between Receptive Fields and Data.}
We want to know if receptive field combinations vary among data.
Therefore, we evaluate the generalization ability of the searched structures
on the subsets of the same dataset and different datasets, respectively.
Within the BreakFast dataset, we perform the global-to-local search on one fold and then
evaluate the searched structures on other folds.
\tabref{tab:ablation_cross_val_splits} shows almost no obvious performance gap on different folds,
indicating that receptive field combinations almost have no difference within a dataset.
However, when searching and evaluating structures across different datasets,
different structures searched on different datasets have a large performance gap, as shown in~\tabref{tab:ablation_cross_val_datasets}.
We can conclude that different data distributions will result in different receptive field combinations.
We visualize the structures searched from different datasets in~\figref{fig:architectures_datasets}.
The searched structures are based on both global and local search. Since the global search introduces the randomness
of each structure, we cannot fairly compare these structures for different datasets.
Still, we give a rough explanation based on the searched receptive field of each structure.
The structures searched on the Breakfast and 50Salads datasets tend to have larger receptive fields, while the structure searched on the GTEA dataset
has smaller receptive fields.
The number of video frames shown in Table 2 positively correlates with receptive fields.
We assume that the average video length of a dataset might influence the receptive fields of the structure.
A video with more frames normally
requires larger receptive fields to capture long-term relations.
The structures searched on Breakfast and 50Salads datasets have similar average receptive fields,
but the average video length of 50Salads is longer than Breakfast.
We assume that understanding the content of cooking breakfast in the Breakfast dataset
requires more long-range information than
the preparing salads content in the 50Salads dataset.
Analyzing different video contents might need features from different ranges of receptive fields.
Our work mainly focuses on searching receptive fields,
but fully explaining why the receptive field combination of a certain task on a certain dataset looks like the searched one is still an open question.

\myPara{Receptive Fields for Different Stages.}
Our global-to-local search is based on MS-TCN.
MS-TCN contains four stages, and all stages share the same receptive field combination in human design.
The visualized searched structures shown in~\figref{fig:architectures_datasets} demonstrate that
different stages have different receptive field combinations,
which conflicts with human design.
We further count the average receptive fields of each stage among all individuals. The range of performance and the average dilation rates of each stage are shown in~\figref{fig:arverage_dilation}.
The average dilation rate in the first stage of MS-TCN tends to be large on high-performance structures.
In contrast, the average dilation rate in the third stage of MS-TCN is relatively small on high-performance structures.
We assume that the first stage of MS-TCN requires large receptive fields to
get the long-term context for coarse prediction, while the following stages need small receptive fields to refine the results locally.

\begin{figure*}[!t]
	\centering
	\small
	\begin{overpic}[width=\linewidth]{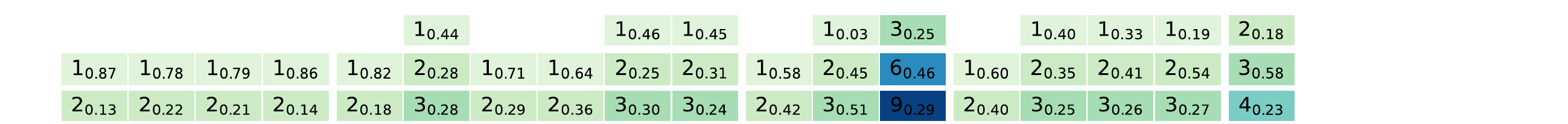}
	\end{overpic}
	\begin{overpic}[width=\linewidth]{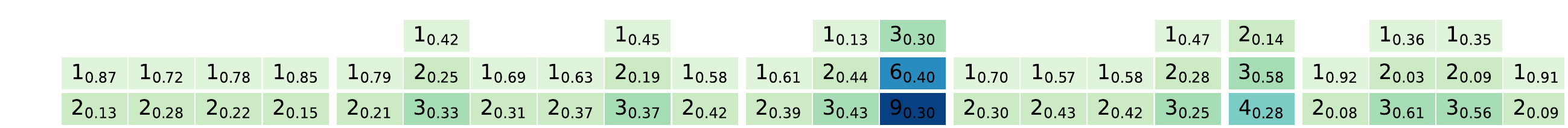}
		\put(12.0, -1.7){S2}
		\put(33.0, -1.7){S3}
		\put(53.2, -1.7){S4}
		\put(68.0, -1.7){FPN}
		\put(78.5, -1.7){RPN}
		\put(89.0, -1.7){MASK}

		\put(0, 10){(a)}
		\put(0, 2){(b)}
	\end{overpic}
	\caption{Visualization and probability of each receptive field in the parallel searched structure
		of ResNet-50 based Faster-RCNN (a) and Mask-RCNN (b) when $S$=3 during searching.
		S2-S4 denotes stage 2 to stage 4 of the ResNet backbone.
		FPN, RPN, and MASK mean the feature pyramid network, region proposal network, and the mask segmentation head
		in Faster-RCNN and Mask-RCNN.}
	\label{fig:architectures_faster_r50_multi}
\end{figure*}

\begin{table*}[t]
	\small
	\centering
	\caption{Performance of local search on instance segmentation with COCO~\cite{lin2014microsoft} dataset
		using Mask-RCNN as the baseline method.
		Local-P means the local-searched structure with parallel receptive fields as described in~\secref{sec:ls}.
		-R50 and -R101 denote using ResNet-50 and ResNet-101 as backbones, respectively.
		R50-NonLocal means adding the non-local block~\cite{wang2018non} to each residual block in stage 4 of the ResNet-50 backbone.
		$S$ indicates using $S$ branches in local search as shown in~\eqref{equ:normalize}.}
	\setlength{\tabcolsep}{1.0mm}
	\begin{tabular}{lccccccc|cccccc}  \toprule
		                                & P              & mAP$^{\rm bb}$      & mAP$^{\rm bb}_{50}$ & mAP$^{\rm bb}_{75}$    & mAP$^{\rm bb}_{\rm s}$ & mAP$^{\rm bb}_{\rm m}$ & mAP$^{\rm bb}_{\rm l}$
		                                & mAP$^{\rm mk}$ & mAP$^{\rm mk}_{50}$ & mAP$^{\rm mk}_{75}$ & mAP$^{\rm mk}_{\rm s}$ & mAP$^{\rm mk}_{\rm m}$ & mAP$^{\rm mk}_{\rm l}$                                                                    \\
		\midrule
		\multicolumn{13}{l}{validation set:}                                                                                                                                                                                                       \\
		\midrule
		Mask-RCNN-R50~\cite{He_2017}    &                & 38.2                & 59.0                & 41.7                   & 22.4                   & 41.6                   & 49.2                   &
		34.7                            & 55.9           & 37.0                & 16.5                & 37.4                   & 50.1                                                                                                               \\
		+RF ($ S=3$)                    &                & 39.7                & 60.7                & 43.4                   & 23.7                   & 43.3                   & 51.3                   &
		36.0                            & 57.8           & 38.4                & 17.0                & 39.1                   & 52.2                                                                                                               \\
		+RF ($ S=3$)                    & \checkmark     & 41.0                & 62.3                & 44.9                   & 24.2                   & 44.7                   & 53.2                   &
		37.1                            & 59.1           & 39.5                & 18.4                & 39.9                   & 53.6                                                                                                               \\
		+RF ($ S=2$)                    &                & 39.6                & 60.8                & 43.2                   & 23.3                   & 43.1                   & 51.4                   &
		35.9                            & 57.6           & 38.3                & 16.4                & 38.9                   & 52.2                                                                                                               \\
		+RF ($ S=2$)                    & \checkmark     & 40.7                & 62.1                & 44.5                   & 24.4                   & 44.3                   & 52.9                   &
		36.8                            & 58.9           & 39.3                & 18.0                & 39.7                   & 53.6                                                                                                               \\
		\midrule
		Mask-RCNN-R101~\cite{He_2017}   &                & 40.0                & 60.5                & 44.0                   & 22.6                   & 44.0                   & 52.6                   &
		36.1                            & 57.5           & 38.6                & 18.8                & 39.7                   & 49.5                                                                                                               \\
		+RF ($ S=3$)                    &                & 41.7                & 62.8                & 45.7                   & 24.8                   & 45.4                   & 55.4                   &
		37.4                            & 59.6           & 40.0                & 18.5                & 40.6                   & 55.1                                                                                                               \\
		+RF ($ S=3$)                    & \checkmark     & 42.7                & 64.0                & 46.7                   & 25.6                   & 46.8                   & 55.8                   &
		38.4                            & 60.9           & 41.3                & 18.5                & 42.0                   & 55.7                                                                                                               \\
		\midrule
		R50-NonLocal~\cite{wang2018non} &                & 39.4                & 60.8                & 43.4                   & 23.2                   & 43.2                   & 50.5                   & 35.5 & 57.1 & 37.9 & 17.2 & 38.5 & 51.3 \\
		+RF ($ S=2$)                    &                & 40.2                & 61.5                & 43.9                   & 23.9                   & 43.9                   & 52.2                   & 36.4 & 58.5 & 38.7 & 17.6 & 39.3 & 52.3 \\
		+RF ($ S=2$)                    & \checkmark     & 40.8                & 62.2                & 44.6                   & 24.4                   & 44.5                   & 53.0                   & 36.9 & 59.0 & 39.6 & 18.0 & 39.8 & 53.8 \\
		\midrule
		\multicolumn{13}{l}{test set:}                                                                                                                                                                                                             \\
		\midrule
		Mask-RCNN-R50~\cite{He_2017}    &                & 38.5                & 59.5                & 41.8                   & 22.3                   & 41.6                   & 47.4                   &
		34.9                            & 56.4           & 37.2                & 18.9                & 37.5                   & 44.6                                                                                                               \\
		+RF ($ S=3$)                    &                & 40.0                & 61.4                & 43.6                   & 23.2                   & 42.6                   & 49.8                   &
		36.2                            & 58.5           & 38.5                & 19.9                & 38.6                   & 46.8                                                                                                               \\
		+RF ($ S=3$)                    & \checkmark     & 41.0                & 62.5                & 45.0                   & 24.0                   & 43.8                   & 51.3                   &
		37.1                            & 59.5           & 39.9                & 20.5                & 39.7                   & 48.2                                                                                                               \\

		+RF ($ S=2$)                    &                & 39.8                & 61.2                & 43.4                   & 23.2                   & 42.3                   & 49.4                   &
		36.1                            & 58.2           & 38.6                & 19.8                & 38.4                   & 46.5                                                                                                               \\
		+RF ($ S=2$)                    & \checkmark     & 41.0                & 62.5                & 44.7                   & 24.1                   & 43.7                   & 51.4                   &
		37.1                            & 59.5           & 39.7                & 20.7                & 39.6                   & 48.3                                                                                                               \\
		\midrule
		Mask-RCNN-R101~\cite{He_2017}   &                & 40.4                & 61.2                & 44.1                   & 23.1                   & 43.5                   & 50.8                   &
		36.5                            & 58.3           & 38.9                & 19.6                & 39.2                   & 47.8                                                                                                               \\
		+RF ($ S=3$)                    &                & 42.0                & 63.4                & 45.9                   & 24.2                   & 45.0                   & 53.1                   &
		37.8                            & 60.5           & 40.4                & 20.5                & 40.4                   & 49.7                                                                                                               \\
		+RF ($ S=3$)                    & \checkmark     & 42.8                & 64.1                & 46.9                   & 24.7                   & 46.1                   & 54.2                   &
		38.5                            & 61.3           & 41.3                & 21.0                & 41.4                   & 50.7                                                                                                               \\
		\midrule
		R50-NonLocal~\cite{wang2018non} &                & 39.8                & 61.4                & 43.3                   & 23.2                   & 42.8                   & 49.1                   & 36.0 & 58.2 & 38.3 & 19.7 & 38.6 & 46.3 \\
		+RF ($ S=2$)                    &                & 40.4                & 61.9                & 44.0                   & 23.2                   & 42.8                   & 50.4                   & 36.4 & 58.7 & 38.9 & 19.7 & 38.6 & 47.2 \\
		+RF ($ S=2$)                    & \checkmark     & 41.2                & 62.9                & 45.0                   & 23.7                   & 43.7                   & 51.7                   & 37.3 & 59.8 & 39.9 & 20.3 & 39.7 & 48.3 \\
		\bottomrule
	\end{tabular}
	\label{tab:instance_segmentation}
\end{table*}

\section{RF-Next for Multiple Networks and Tasks}
This section shows that RF-Next models can be applied
to multiple networks and tasks. We use prefixes RF to denote RF-Next models, and P means the parallel receptive field version.

\subsection{Spatial tasks}

We apply our proposed searching scheme to
find proper receptive fields for spatial tasks,
\eg object detection, instance segmentation, and semantic segmentation.
We observe that proper receptive fields significantly improve the performance of these tasks.

\subsubsection{Object Detection}
Object detection aims to assign bounding boxes and categories to objects of various sizes.
We utilize the widely used Faster-RCNN~\cite{fastrrcnn} method with a dilation rate of 1
for all convolutions.
Faster-RCNN applies the feature pyramid network to aggregate features with multiple scales
to handle objects of various sizes.
However, the receptive fields for convolutions are ignored.
Therefore, we search for dilation rates of convolutions with kernel sizes larger than one in the Faster-RCNN.
Due to the extremely large training cost and the small initial dilation rates,
we only use the highly efficient local search scheme.

\myPara{Training and Searching.}
We employ the ResNet-50~\cite{he2016deep} and ResNet-101~\cite{he2016deep} that are pre-trained on the
ImageNet~\cite{imagenet_cvpr09} dataset as backbones.
We verify the effectiveness of local search on the COCO dataset~\cite{lin2014microsoft}
and report the mean average precision (mAP) to evaluate the trained model.
Following the official training scheme~\cite{fastrrcnn},
the images are resized to $1333\times800$ with a randomly horizontal
flip, and the model is trained for 12 epochs, with a batch size of 16 on 4 GPUs.
During the local search, we train the model for 12 epochs
and update the structure in each epoch for the first 10 epochs.
The $\Delta D_l$ is set to be $0.5D_l$.
Since the weights of the first stage of ResNet in Faster-RCNN are frozen during training,
we skip this stage during searching.

\myPara{Performance and Observation.}
As shown in~\tabref{tab:object_detection},
RF-Next model improves the test mAP of Faster-RCNN with ResNet-50 by 1.4\%.
For the RF-ResNet-101 model, the test mAP is also improved by 1.5\%.
Theoretically, a network with a larger depth has a larger range of receptive fields.
Still, effective receptive field settings have a similar performance gain on both shallow and deep models.
We visualize the searched dilation rates of RF-ResNet-50/101 based Faster-RCNN as shown in~\figref{fig:architectures_detection}.
The shallow layers require relatively small dilation rates,
while some deep layers have large dilation rates.
Interestingly, the ResNet-101 based model requires larger dilation rates in stage 4 of the
network than ResNet-50 based model.

When utilizing the RF-Next with parallel receptive fields,
the performance gain of ResNet-50 and ResNet-101 based models are 2.6\% and 2.4\% in test mAP,
indicating that object detection task needs parallel multi-scale ability.
The visualization and probability of each receptive field in the parallel RF-Next
are shown in~\figref{fig:architectures_faster_r50_multi}.
By default, we utilize the number of sampling dilation rates $S=3$ in \eqref{equ:normalize}.
And we also explore using $S=2$ for local search, as shown in~\tabref{tab:object_detection}.
Using two/three branches achieve similar performance for parallel RF-Next, showing that using two branches for each layer
provides sufficient multi-scale ability.
The result is also consistent with
the observation in~\tabref{tab:ablation_s} that the proposed expectation-guided search
is insensitive to the number of sampling dilation rates.

We analyze the performance gain for objects of different sizes in~\tabref{tab:object_detection}.
For ResNet-50 and ResNet-101 based models, the test mAP improvement for small, medium, and large objects
are (1.5\%, 2.2\%, 4.1\%) and (1.8\%, 2.2\%, 3.4\%), respectively.
The performance gain gradually increases with the increase of object sizes,
showing that the default receptive field settings of Faster-RCNN are not large enough to capture large objects.

\subsubsection{Instance Segmentation}
Instance segmentation outputs the instances segmentation masks and categories,
which is similar to the object detection task.
To compare the receptive field requirements of object detection and instance segmentation,
we use the widely used Mask-RCNN~\cite{He_2017} that extends the Faster-RCNN with
a mask segmentation branch.
Like the object detection,
we apply the local search on convolutions whose kernel is larger than one.

\myPara{Training and Searching.}
The ResNet-50~\cite{he2016deep} and ResNet-101~\cite{he2016deep} with ImageNet~\cite{imagenet_cvpr09} pre-training are used as backbones.
We use the COCO dataset~\cite{lin2014microsoft} for object detection
and report the mean average precision of bounding box (mAP$^{\rm bb}$) and
instance segmentation (mAP$^{\rm mk}$) to evaluate the trained model.
To fairly compare the searched structure on instance segmentation and object detection,
the training scheme of Mask-RCNN is aligned with Faster-RCNN in both searching and re-training stages.

\myPara{Performance and Observation.}
We give the performance comparison of searched structure and baseline in~\tabref{tab:instance_segmentation}.
Using the RF-Next with a single branch brings
1.3\%/1.3\% gain on test mAP$^{\rm mk}$ and 1.5\%/1.6\% gain on test mAP$^{\rm bb}$
of ResNet-50/101 based models.
And the performance gains are further enlarged by the parallel RF-Next,
\ie 2.2\%/2.0\% gain on test mAP$^{\rm mk}$ and 2.5\%/2.4\% gain on test mAP$^{\rm bb}$.
We give the visualized dilation rates comparison between Faster-RCNN and Mask-RCNN as shown in~\figref{fig:architectures_detection}.
The dilation rates of these two tasks are very similar because of the large similarity between them.
As shown in~\figref{fig:architectures_faster_r50_multi}, the probability of dilation rates in the mask segmentation head of Mask-RCNN shows that the middle two layers require diverse receptive fields
while the first and last layer requires a small receptive field.

\subsubsection{Semantic Segmentation}
Semantic segmentation task requires assigning each pixel of images
with category labels.
Receptive fields are vital for the dense pixel-level prediction
of semantic segmentation.
Deeplab series~\cite{chen2018deeplab,chen2017rethinking}
utilize convolution with dilation rates larger than one
to enlarge the receptive fields,
which becomes the default option for semantic segmentation networks~\cite{huang2019ccnet,yang2018denseaspp}.
However, possible better receptive fields than the human-designed ones have not been explored.
We apply the Deeplab V3~\cite{chen2017rethinking} network as the baseline method and conduct
local search to find more effective receptive fields.

\myPara{Training and Searching.}
We conduct experiments on both the PASCAL VOC~\cite{Everingham2009ThePV} dataset
and the ADE20k dataset~\cite{Zhou_2017_CVPR} using the ImageNet pretrained ResNet-50~\cite{he2016deep} based Deeplab V3 as the baseline method.
Following official training settings~\cite{chen2017rethinking},
models are trained for 20k and 80k iterations for PASCAL VOC and ADE20k, respectively,
with a batch size of 16 on 4 GPUs.
For both datasets, the images are randomly scaled at a ratio between 0.5 and 2.0
and randomly cropped to $512\times512$.
An auxiliary loss is applied at the output of the third stage of the ResNet backbone
to ease convergence.
To avoid the influence of auxiliary loss,
we apply the local search to stage 4 and the decoder of the network.
During the local search,
we update the structure every 2k and 8k iterations for PASCAL VOC and ADE20k, respectively,
and conduct 8 iterations of the local search in total.
The $\Delta D_l$ is set to be $0.15D_l$ for both datasets.

\begin{figure}[!t]
	\centering
	\small
	\begin{overpic}[width=0.4\linewidth]{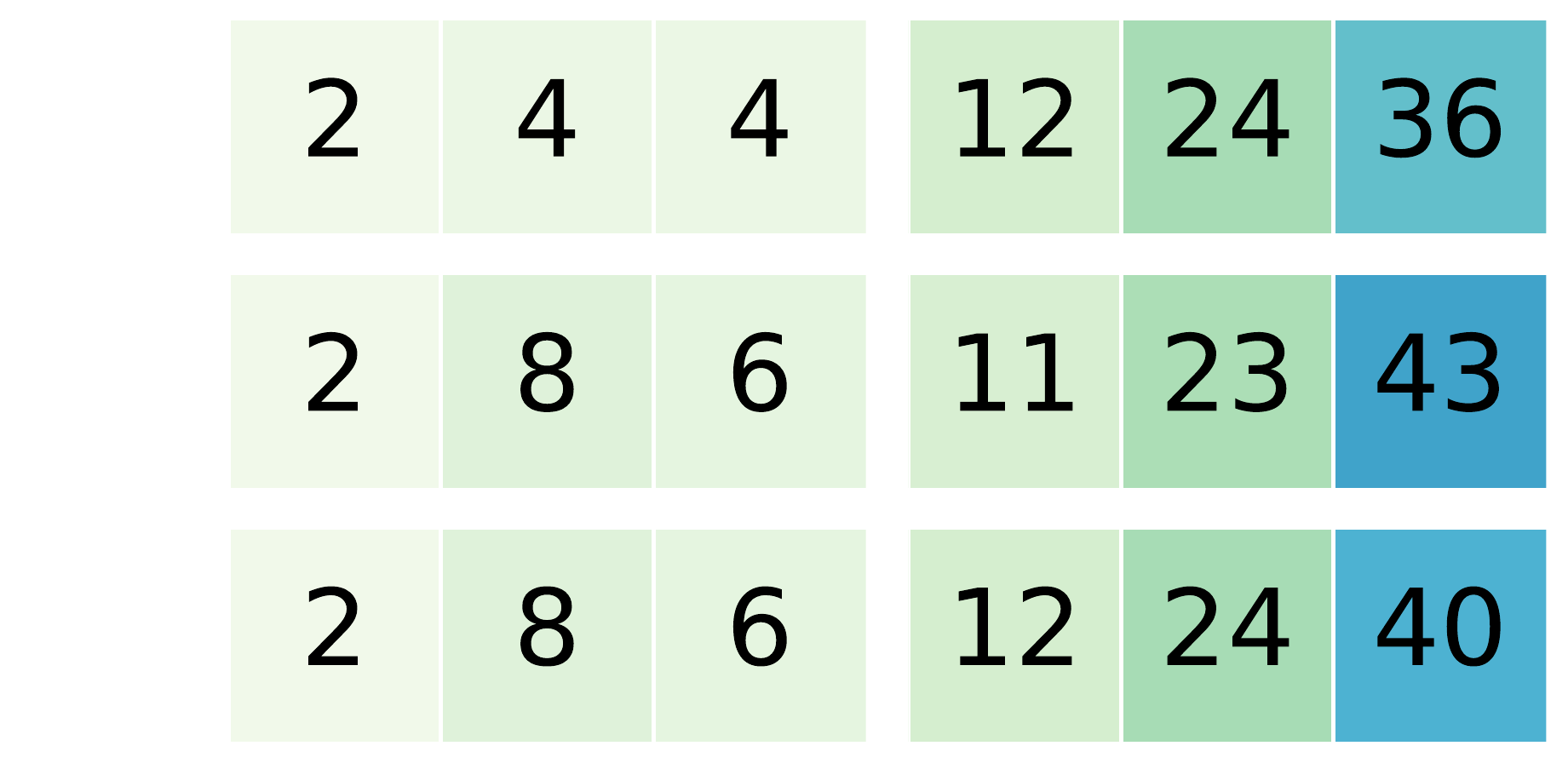}
		\put(-45, 38){Baseline}
		\put(-45, 21.5){PASCAL VOC}
		\put(-45, 5.2){ADE20K}
		\put(30, -6.5){S4}
		\put(64, -6.5){Decoder}
	\end{overpic}
	\caption{Visualization of the local searched receptive fields of
		stage 4 and decoder in Deeplab V3 on the semantic segmentation task.}
	\label{fig:architectures_semantic}
\end{figure}

\myPara{Performance and Observation.}
We utilize the mean intersection over union~(mIoU) and
the mean accuracy~(mAcc) to evaluate the trained models.
As shown in~\tabref{tab:semantic_seg},
the RF-Next brings the 1.6\% and 0.8\% gain in mIoU for PASCAL VOC and
ADE20k datasets, respectively.
The searched multi-branch structures achieve similar performance to the single branch competitors.
We assume the atrous spatial pyramid pooling structure in Deeplab V3
already enhances the parallel multi-scale ability of the network.
The visualization of searched receptive fields in~\figref{fig:architectures_semantic}
indicates that larger receptive fields are required in stage 4 of the network.

As described above, we skip the local search in the first three stages of the network
to avoid the side effect of auxiliary loss.
We show in~\tabref{tab:semantic_seg} that searching all convolutions achieves 76.3\% mIoU,
similar to the human-designed baseline.
We observe that searching receptive fields for all convolutions results
in a much smaller auxiliary loss value than human-designed baseline (0.102 VS. 0.191)
and searching after stage 3 (0.102 VS. 0.189).
Also, compared to searching after stage 3,
searching in the early stages achieves the performance gain of 15.4\% in mIoU
for output in stage 3 of the network.
Since the local search relies on the gradient backpropagation to find receptive fields,
adding auxiliary loss makes the local search find better receptive fields for output in stage 3
instead of the final output.

\begin{table}[t]
	\small
	\centering
	\caption{Performance of local search on semantic segmentation
		with PASCAL VOC~\cite{Everingham2009ThePV} and ADE20k datasets~\cite{Zhou_2017_CVPR}.
		Local-P means the local-searched structure with parallel receptive fields as described in~\secref{sec:ls}.
		\dag means apply local search to all convolutions of the network.
		Local-S3 indicates the output in stage 3 of the network,
		where an auxiliary loss is added to ease convergence~\cite{chen2017rethinking}.
	}
	\setlength{\tabcolsep}{2.5mm}
	\begin{tabular}{lccccc}  \toprule
		                                    & \multirow{2}{*}{P} & \multicolumn{2}{c}{VOC~\cite{Everingham2009ThePV}} & \multicolumn{2}{c}{ADE20K~\cite{Zhou_2017_CVPR}}               \\
		                                    &                    & mIoU                                               & mAcc                                             & mIoU & mAcc \\
		\midrule
		DeepLabV3~\cite{chen2017rethinking} &                    & 76.2                                               & 85.7                                             & 42.4 & 53.6 \\
		+RF                                 &                    & 77.8                                               & 87.4                                             & 43.2 & 54.1 \\
		+RF                                 & \checkmark         & 77.9                                               & 87.6                                             & 43.0 & 53.7 \\ \midrule
		+RF\dag                             &                    & 76.3                                               & 85.8                                             & -    & -    \\
		+RF-S3                              &                    & 51.7                                               & 64.7                                             & -    & -    \\
		+RF-S3\dag                          &                    & 67.7                                               & 79.8                                             & -    & -    \\
		\bottomrule
	\end{tabular}
	\label{tab:semantic_seg}
\end{table}

\subsection{Sequential tasks}
Except for the temporal action segmentation task,
we also conduct receptive field search on other sequential tasks, \eg speech synthesis,
P-MNIST digit classification, and polyphonic music modeling.
We verify the effectiveness of global-to-local search on the polyphonic music modeling
and P-MNIST digit classification task.
Due to the high training cost of speech synthesis, we apply the local search on top of the human-designed structure.

\subsubsection{Speech Synthesis}
As described in~\secref{sec:tts_related},
we focus on the procedure of transferring acoustic features to speech waveform in speech synthesis.
We use WaveGlow~\cite{8683143} as the baseline method
that combines the advantage of Glow~\cite{NEURIPS2018_d139db6a} and WaveNet~\cite{oord2016wavenet}.
WaveGlow network has 12 coupling layers, where each contains 8 layers of dilated convolutions
with human-designed gradually expanded dilation rates.
To save computational cost,
we utilize the local search to find more effective dilation rates
of these layers based on human-designed structures.

\myPara{Training and Searching.}
We conduct experiments on the widely used LJ speech~\cite{ljspeech17} dataset,
including 13,100 audio clips with a total length of about 24 hours.
Each sample is randomly cropped to 16,000 for training, and the sampling rate is 22,050Hz.
The mel-spectrograms~\cite{Wang2017,shen2018natural},
generated through a short-time Fourier transform,
are fed to the network for speech synthesis.
The network is optimized by Adam optimizer with a learning rate of 1e-4 for 100 epochs.
We train the model with 4 GPUs using the batch sizes of 20 and 48 during searching and training, respectively.
During the local searching, the $\Delta D_l$ and $S$ are set to be $0.6D_l$ and 3, respectively.
We train the model for 60 epochs during the local search and update the structure every 3 epochs.

\myPara{Performance and Observation.}
We use three metrics, \ie mel-cepstral distortion (MCD)~\cite{407206},
perceptual evaluation of speech quality (PESQ)~\cite{941023}
and log-likelihood ratio (LLR)~\cite{quackenbush1988objective},
to evaluate the speech synthesis quality, as shown in~\tabref{tab:WaveGlow}.
MCD measures the difference between two sequences of mel-cepstra of speech,
and a small MCD indicates the synthesized and natural speeches are close.
Similarly, LLR measures the difference between two speeches.
The structure with searched receptive field combinations achieves better performance
than human-designed receptive fields
in MCD and PESQ.
PESQ assesses the voice quality, and a higher value
between the synthesized and natural speech means better the synthesized speech quality.
We calculated the PESQ under the narrowband of 8,000Hz.
The speech synthesis results of the local-searched structure also have a better PESQ score,
indicating that more proper receptive fields
benefit the speech synthesis quality.

\begin{figure*}[!t]
	\centering
	\small
	\begin{overpic}[width=\linewidth]{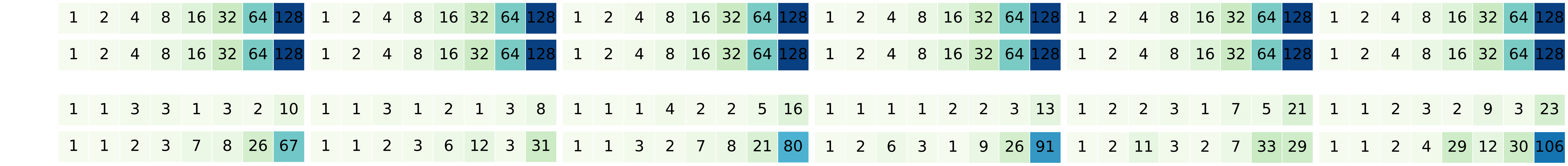}
		\put(0, 6.5){\rotatebox{90}{Base}}
		\put(0, 0.7){\rotatebox{90}{Ours}}
	\end{overpic}
	\caption{Visualization of the local searched dilation rates of WaveGlow~\cite{8683143}
		for the speech synthesis task. }
	\label{fig:architectures_tts}
\end{figure*}

We visualize the local searched and baseline receptive fields (dilation rates) of WaveGlow~\cite{8683143} in~\figref{fig:architectures_tts}.
We observe that the maximum dilation rate of the human-designed structure is much larger
than the searched structure, indicating that too large receptive fields may not be necessary for this task.
Unlike the human-designed structure with the same receptive field combination for each coupling layer,
the searched structure has small receptive fields on the shallow layers and larger receptive fields
on the deeper layers.
We assume that the speech synthesis task requires local features in the shallow layers,
and the deeper layers are responsible for modeling the long-term dependencies.

\begin{figure}[!t]
	\centering
	\small
	\begin{overpic}[width=0.4\linewidth]{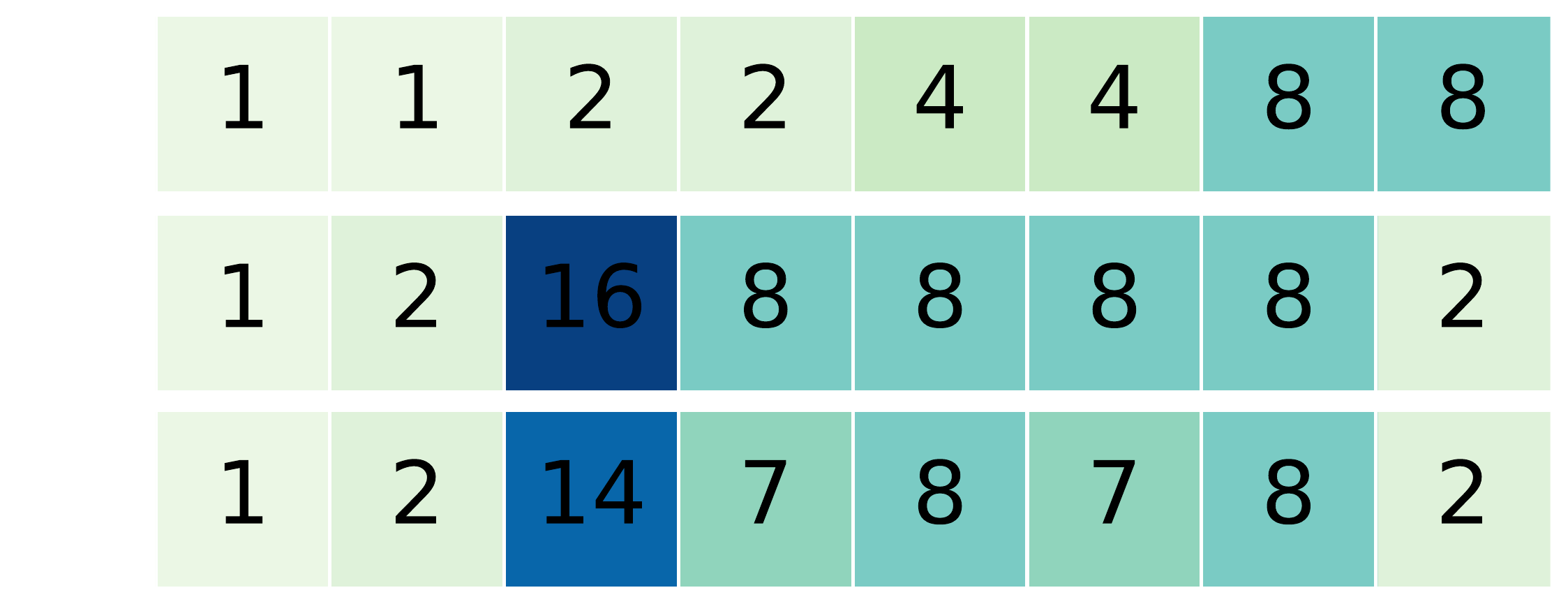}
		\put(-10, 29){(a)}
		\put(-10, 17){(b)}
		\put(-10, 5){(c)}
	\end{overpic}
	\caption{Visualization of the baseline structure~(a),
		the global searched structure~(b) and
		the global-to-local searched structure~(c)
		of TCN on polyphonic music modeling task.}
	\label{fig:architectures_music}
\end{figure}
\begin{figure}[!t]
	\centering
	\small
	\begin{overpic}[width=0.9\linewidth]{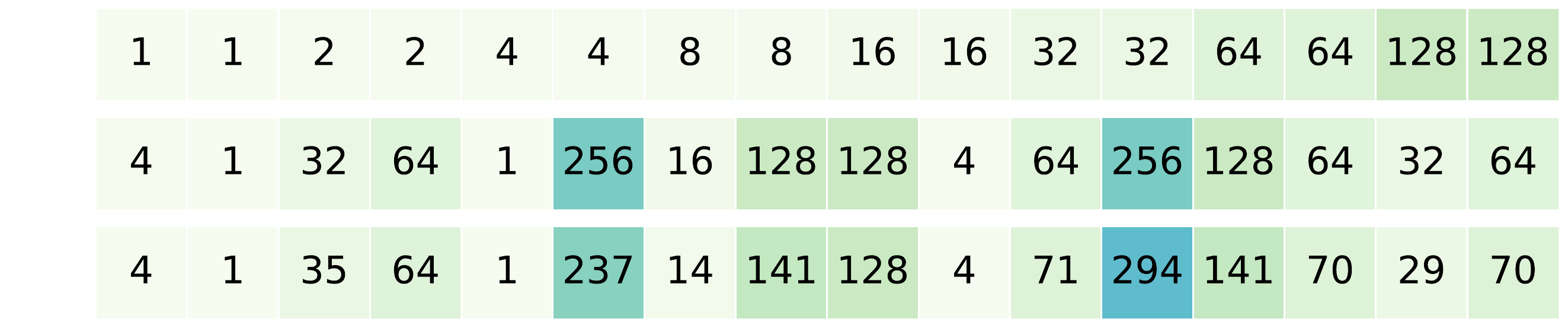}
		\put(0, 16.6){(a)}
		\put(0, 9.6){(b)}
		\put(0, 2.6){(c)}
	\end{overpic}
	\caption{Visualization of the baseline structure~(a),
		the global searched structure~(b) and
		the global-to-local searched structure~(c)
		of TCN on P-MNIST Classification task.}
	\label{fig:architectures_pmnist}
\end{figure}

\begin{table}[t]
	\centering
	\caption{Speech synthesis performance of WaveGlow~\cite{8683143} based on local-searched structure on the LJ speech dataset~\cite{ljspeech17}.}
	\setlength{\tabcolsep}{2.9mm}
	\begin{tabular}{lcccc}  \toprule
		                        & MCD$\downarrow$~\cite{407206} & LLR$\downarrow$~\cite{quackenbush1988objective} & PESQ$\uparrow$~\cite{941023} \\
		\midrule
		WaveGlow~\cite{8683143} & 5.79                          & 1.29                                            & 1.52                         \\
		RF-WaveGlow             & 5.59                          & 0.71                                            & 1.84                         \\
		\bottomrule
	\end{tabular}
	\label{tab:WaveGlow}
\end{table}

\subsubsection{Sequence Modeling using TCN}
Bai \etal~\cite{BaiTCN2018} have verified the performance of TCN on several sequence modeling tasks.
We further show the effectiveness of RF-Next models on two sequential tasks,
\ie polyphonic music modeling and P-MNIST digit classification.

\myPara{P-MNIST Classification.}
P-MNIST classification aims to classify the handwritten digit images
in which the order of the pixels is disrupted.
P-MNIST dataset~\cite{le2015simple,zhang2016architectural} randomly permutes images in the MNIST dataset~\cite{lecun1998gradient} to 784 length sequences,
which is widely used for long-term relation modeling~\cite{wisdom2016full,cooijmans2016recurrent,krueger2016zoneout,jing2017tunable,BaiTCN2018}.
A TCN with 8 layers is used for P-MNIST classification~\cite{BaiTCN2018},
where each layer has a convolution with a kernel size of 7 and a channel number of 25.
We apply the global-to-local search on the TCN to find more effective receptive fields.
Following the settings in~\cite{BaiTCN2018},
the final model is trained for 20 epochs with an Adam optimizer.
The initial learning rate is 2e-3 and is multiplied by 0.1 at the 10-th epoch.
In P-MNIST, the order of the pixels is randomly permuted,
and we fix the permutation order for all experiments.
During the global search, we set the iterations $N=50$ and
the initial population size $M=25$.
The model is trained with 11 epochs for each sample.
During the local search, the $\Delta D_l$ is set to $0.1D_l$,
and the structure is trained for 15 epochs and updated every 3 epochs.
The classification accuracy is used as the evaluation metric.
As shown in~\tabref{tab:tcn},
the global search improves the accuracy from 97.2\% to 97.6\%,
and the local search further improves the performance to 97.8\%.
The visualized structure in~\figref{fig:architectures_pmnist} shows the searched receptive
fields are very different from the human-designed patterns.

\myPara{Polyphonic Music Modeling.}
Polyphonic music modeling aims to predict the subsequent musical
notes based on the history of the played notes.
The polyphonic music modeling is conducted on the widely used Nottingham~\cite{chung2014empirical,pascanu2013construct,jozefowicz2015empirical} dataset
composed of 1200 British and American folk tunes.
For polyphonic music modeling, we utilize a TCN with 4 layers,
where each layer has two convolutions with a kernel size of 5 and a channel number of 150.
Following~\cite{BaiTCN2018}, the model is trained for 100 epochs with an Adam optimizer.
The initial learning rate is 1e-3, which is multiplied by 0.1 every 30 epochs.
The dropout with the rate of 0.25 and the gradient clipping with the maximum norm of 0.2
is applied.
We set the iterations $N=50$ and the initial population size $M=25$ for the global search,
and each sample is trained with 30 epochs.
For local search, the $\Delta D_l$ is set to $0.15D_l$,
and the model is trained for 60 epochs and updated in every 10 epochs.
We evaluate models using the negative-log-likelihood~(NLL),
as shown in~\tabref{tab:tcn}.
The NLL is improved from 2.97 to 2.73 by the global search,
and the local search improves the performance to 2.69.
The global and local-searched structures
are shown in \figref{fig:architectures_music}.

\begin{table}[t]
	\small
	\centering
	\caption{Performance of global-to-local search on TCN~\cite{BaiTCN2018}.
		We evaluate the performance of the polyphonic music modeling and P-MNIST digit classification task.}
	\setlength{\tabcolsep}{1.9mm}
	\begin{tabular}{lcccc}  \toprule
		Task                                & baseline & global & global+local \\
		\midrule
		Permuted MNIST~(accuracy$\uparrow$) & 97.2     & 97.6   & 97.8         \\
		Music Nottingham~(NLL$\downarrow$)  & 2.97     & 2.73   & 2.69         \\
		\bottomrule
	\end{tabular}
	\label{tab:tcn}
\end{table}

\subsection{Receptive field search on Modern Networks}
We apply the receptive field searching method
on multiple networks, \eg SOTA attention/convolution based networks, multi-scale networks, and searched networks.
\begin{table}[t]
	\small
	\centering
	\caption{Receptive field search improves SOTA attention/convolution models for object detection and instance segmentation tasks on the COCO val dataset.
		Following the official implementation of PVT~\cite{wang2021pyramid,wang2021pvtv2} and ConvNeXt~\cite{liu2022convnet}, PVTv2-B0 and ConvNeXt-T adopt the Mask RCNN detector and Cascade Mask RCNN detector, respectively.
	}
	\setlength{\tabcolsep}{0.8mm}
	\begin{tabular}{lccccccc}  \toprule
		Object det.   & P          & mAP  & mAP$_{50}$ & mAP$_{75}$ & mAP$_{\rm s}$ & mAP$_{\rm m}$ & mAP$_{\rm l}$ \\
		\midrule
		PVTv2-B0      &            & 38.2 & 60.5       & 40.7       & 22.9          & 40.9          & 49.6          \\
		RF-PVT        &            & 38.8 & 60.9       & 41.8       & 23.6          & 41.2          & 50.8          \\
		RF-PVT        & \checkmark & 39.1 & 60.8       & 42.7       & 23.3          & 41.8          & 51.4          \\
		\midrule
		ConvNeXt-T    &            & 50.4 & 69.1       & 54.8       & 33.9          & 54.5          & 65.1          \\
		RF-ConvNeXt   &            & 50.6 & 69.2       & 54.8       & 34.1          & 54.0          & 65.5          \\
		RF-ConvNeXt   & \checkmark & 50.9 & 69.5       & 55.5       & 34.3          & 54.6          & 65.8          \\  \midrule
		Instance seg. &            & mAP  & mAP$_{50}$ & mAP$_{75}$ & mAP$_{\rm s}$ & mAP$_{\rm m}$ & mAP$_{\rm l}$ \\
		\midrule
		PVTv2-B0      &            & 36.2 & 57.8       & 38.6       & 18.0          & 38.4          & 51.9          \\
		RF-PVT        &            & 36.8 & 58.4       & 39.5       & 18.7          & 39.0          & 52.7          \\
		RF-PVT        & \checkmark & 37.1 & 58.5       & 40.0       & 17.8          & 39.3          & 53.7          \\
		\midrule
		ConvNeXt-T    &            & 43.7 & 66.5       & 47.3       & 24.2          & 47.1          & 62.1          \\
		RF-ConvNeXt   &            & 44.0 & 66.8       & 47.5       & 24.8          & 47.0          & 62.1          \\
		RF-ConvNeXt   & \checkmark & 44.3 & 67.3       & 47.8       & 24.7          & 47.4          & 62.6          \\
		\bottomrule
	\end{tabular}
	\label{tab:pvt_det}
\end{table}

\begin{table}[t]
	\small
	\centering
	\caption{Performance of receptive field search on semantic segmentation using PVTv2-B0 backbone with the Semantic FPN method~\cite{wang2021pvtv2}.}
	\setlength{\tabcolsep}{3.4mm}
	\begin{tabular}{lccccc}  \toprule
		         & \multirow{2}{*}{P} & \multicolumn{2}{c}{Pascal VOC~\cite{Everingham2009ThePV}} & \multicolumn{2}{c}{ADE20K~\cite{Zhou_2017_CVPR}}               \\
		         &                    & mIoU                                                      & mAcc                                             & mIoU & mAcc \\
		\midrule
		PVTv2-B0 &                    & 73.7                                                      & 85.0                                             & 37.5 & 48.3 \\
		RF-PVT   &                    & 74.4                                                      & 86.0                                             & 38.0 & 48.6 \\
		RF-PVT   & \checkmark         & 74.4                                                      & 85.9                                             & 37.8 & 48.7 \\
		\bottomrule
	\end{tabular}
	\label{tab:semantic_seg_pvt}
\end{table}

\myPara{Receptive Field Search Improves SOTA Models.}
We show recent SOTA models, \eg PVT~\cite{wang2021pyramid,wang2021pvtv2}
and ConvNeXt~\cite{liu2022convnet},
still benefit from our receptive field searching method.
PVTv2~\cite{wang2021pyramid,wang2021pvtv2} is a
self-attention based pyramid vision transformer,
which uses both global self-attentions and depth-wise convolutions.
We apply the receptive field search to convolutions of PVTv2
for object detection, instance segmentation, and semantic segmentation tasks,
achieving stable improvements over the strong PVTv2-B0 baseline.
As shown in~\tabref{tab:pvt_det},
the single-branch RF-PVTv2 has the gain of 0.6\% box mAP and 0.6\%
mask mAP for object detection and instance segmentation tasks.
The parallel RF-PVTv2 further improves 0.3\% box mAP and 0.3\% mask mAP
for these two tasks.
\tabref{tab:semantic_seg_pvt} shows the searched single-branch structure improves
the baselines with 0.7\% and 0.5\% on Pascal VOC and ADE20K datasets.
ConvNeXt~\cite{liu2022convnet} is a SOTA convolutional model
that outperforms many SOTA attention-based models.
Even though ConvNeXt manually tunes the kernel size of convolutions
to support a larger range of receptive fields,
the receptive field search still further improves
the performance on object detection and instance segmentation.
As shown in~\tabref{tab:pvt_det},
the parallel version of searched structure has considerable
gains of 0.5\% box mAP and 0.6\% mask mAP over the strong
ConvNeXt-T model.
The performance gains over these two strong SOTA models
prove the effectiveness of our receptive field searching method.

\begin{table}[t]
	\small
	\centering
	\caption{Receptive field search improves hand-crafted multiple receptive fields models, \ie Res2Net~\cite{pami20Res2net} and HRNet~\cite{SunXLW19,WangSCJDZLMTWLX19}, for object detection
		and instance segmentation tasks on the COCO val dataset.
		The Cascade Mask RCNN method is used as the detector.
	}
	\setlength{\tabcolsep}{0.8mm}
	\begin{tabular}{lccccccc}  \toprule
		Object det.   & P          & mAP  & mAP$_{50}$ & mAP$_{75}$ & mAP$_{\rm s}$ & mAP$_{\rm m}$ & mAP$_{\rm l}$ \\
		\midrule
		Res2Net-101   &            & 46.3 & 64.4       & 50.5       & 27.2          & 50.3          & 60.5          \\
		RF-Res2Net    &            & 46.9 & 65.8       & 51.2       & 28.4          & 50.7          & 62.1          \\
		RF-Res2Net    & \checkmark & 47.9 & 66.6       & 52.2       & 29.7          & 51.9          & 62.8          \\
		\midrule
		HRNetV2p-W18  &            & 41.6 & 58.7       & 45.4       & 23.5          & 44.7          & 54.9          \\
		RF-HRNet      &            & 42.9 & 60.8       & 46.7       & 25.9          & 46.2          & 54.8          \\
		RF-HRNet      & \checkmark & 43.7 & 61.9       & 47.7       & 26.5          & 47.3          & 56.7          \\
		\midrule
		Instance seg. &            & mAP  & mAP$_{50}$ & mAP$_{75}$ & mAP$_{\rm s}$ & mAP$_{\rm m}$ & mAP$_{\rm l}$ \\
		\midrule
		Res2Net-101   &            & 40.0 & 61.7       & 43.3       & 22.2          & 43.8          & 54.1          \\
		RF-Res2Net    &            & 40.7 & 63.2       & 43.9       & 20.4          & 44.0          & 59.0          \\
		RF-Res2Net    & \checkmark & 41.5 & 64.0       & 44.9       & 21.3          & 44.6          & 59.5          \\
		\midrule
		HRNetV2p-W18  &            & 36.4 & 56.3       & 39.3       & 19.1          & 39.1          & 49.5          \\
		RF-HRNet      &            & 37.6 & 58.3       & 40.4       & 19.0          & 40.2          & 53.9          \\
		RF-HRNet      & \checkmark & 38.1 & 59.3       & 41.0       & 19.4          & 40.7          & 55.3          \\
		\bottomrule
	\end{tabular}
	\label{tab:multiscale}
\end{table}

\myPara{Receptive Field Search Improves Multi-scale Models.}
We show the advantage of receptive field searching over popular hand-crafted multi-scale models with
multiple receptive fields, \eg HRNet~\cite{SunXLW19,WangSCJDZLMTWLX19} and Res2Net~\cite{pami20Res2net}.
HRNet parallel processes features of multiple resolutions
to form the multi-scale representation.
Res2Net constructs hierarchical residual-like connections
within a block to enable fine-grained multiple receptive fields.
Despite their good hand-crafted multi-scale ability,
we show in~\tabref{tab:multiscale}
that searched receptive fields constantly improve
their performance on object detection and instance segmentation tasks.
For HRNet, the single/multiple-branch RF-HRNet
improves 1.3\%/2.1\% box mAP on object detection
and 1.2\%/1.7\% mask mAP on instance segmentation.
The single/multiple-branch RF-Res2Net also
improves the Res2Net with 0.6\%/1.6\% box mAP on object
detection and 0.7\%/1.5\% mask mAP on instance segmentation.
Therefore, our receptive field searching method
can further improve the hand-crafted multi-scale models
with better receptive field combinations.

\myPara{Comparison with Attention Mechanisms.}
As discussed in the related work, attention mechanisms
theoretically can form arbitrary receptive fields~\cite{vaswani2017attention,wang2018non,chen2019graph}.
However, we are not aware of the actual receptive field representation ability of attention mechanisms.
Therefore, we propose to compare our receptive field search method with the non-local module~\cite{wang2018non}
on the instance segmentation task.
Following the implementation in~\cite{wang2018non}, we insert the non-local blocks to each residual block in stage 4 of the ResNet50 backbone.
As shown in~\tabref{tab:instance_segmentation}, on the COCO testing set,
the non-local based Mask-RCNN improves the Mask-RCNN baseline with 1.1 mask mAP and 1.3 box mAP.
The Mask-RCNN with searched parallel receptive fields outperforms non-local based Mask-RCNN by 1.1 mask mAP 1.2 box mAP,
showing that the searched receptive fields provide better representation than non-local modules.
Therefore, though the non-local module improves the model performance,
it cannot provide effective receptive fields as strong as our searched receptive fields.
Then, we apply the local search to the non-local based network,
and the hyper-parameters for searching are kept consistent with the standard ResNet.
We observe a further performance gain with both searched single-branch receptive fields and parallel-branch receptive fields.
The single-branch version has the gain of 0.4 mask mAP and 0.6 box mAP, and
the parallel-branch version has the gain of 1.3 mask mAP and 1.4 box mAP.
The receptive field searching scheme can further improve the performance of the non-local based model,
indicating that the non-local module may not be able to cover all effective receptive fields even with its dense connections among pixels.

\myPara{Searching Receptive Fields over the Searched Network.}
Auto-deeplab~\cite{liu2019auto} searches the feature resolution for different stages of the semantic segmentation network.
To verify if it is possible to adjust the receptive fields of Auto-deeplab further,
we conduct the local search on top of the Auto-deeplab.
We follow the implementation of Auto-deeplab~\cite{liu2019auto} to search on the Cityscapes~\cite{Cordts2016Cityscapes} dataset.
As shown in~\tabref{tab:semantic_seg_autodeeplab},
both single/multiple-branch RF-Auto-deeplab brings performance gain over Auto-deeplab in mIoU.
Therefore, our receptive field searching scheme can further benefit the semantic segmentation model with searched feature resolutions.

\begin{table}[t]
	\small
	\centering
	\caption{Performance of local search on semantic segmentation
		with Auto-deeplab~\cite{liu2019auto} and Cityscapes datasets~\cite{Cordts2016Cityscapes}.
		Local-P means the local-searched structure with parallel receptive fields as described in~\secref{sec:ls}.
	}
	\setlength{\tabcolsep}{3mm}
	\begin{tabular}{lcccc}
		\toprule
		                                & P          & mIoU & mAcc \\
		\midrule
		Auto-deeplab~\cite{liu2019auto} &            & 76.0 & 83.7 \\
		RF-Auto-deeplab                 &            & 76.3 & 84.2 \\
		RF-Auto-deeplab                 & \checkmark & 76.7 & 84.3 \\
		\bottomrule
	\end{tabular}
	\label{tab:semantic_seg_autodeeplab}
\end{table}

\section{Conclusion}
We propose a global-to-local search scheme to search for effective receptive
field combinations at a coarse-to-fine scheme.
The global search discovers effective receptive field combinations with better performance than hand designings
but completely different patterns.
The expectation-guided iterative local search scheme enables searching
fine-grained receptive field combinations in the dense search space.
RF-Next models, enhanced with receptive field search scheme, can be plugged into multiple tasks,
\eg action segmentation, sequence modeling~\cite{chen2020probabilistic,BaiTCN2018},
segmentation~\cite{chen2018deeplab,GaoEccv20Sal100K,VecRoad_20CVPR,gu2020pyramid,He_2017},
object detection~\cite{ren2015faster,HanEccv20SemLine} methods to boost the performance further.


\myPara{Acknowledgement}

This research was supported by the Major Project for New Generation of AI
under Grant No. 2018AAA0100400, NSFC (61922046),
and S\&T innovation project from Chinese Ministry of Education.


\bibliographystyle{IEEEtran}
\bibliography{ref}

\newcommand{\AddPhoto}[1]{{\includegraphics[width=1in,keepaspectratio]{fig/authors/#1}}}
\newcommand{\AuthorBio}[3]{\begin{IEEEbiography}[\AddPhoto{#1}]{#2}#3\end{IEEEbiography}\vspace{-.2in}}

\AuthorBio{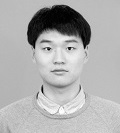}{Shanghua Gao}{
	is a Ph.D. candidate in Media Computing Lab at Nankai University.
	He is supervised via Prof. Ming-Ming Cheng.
	His research interests include computer vision and representation learning.
}

\AuthorBio{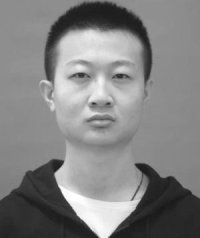}{Zhong-Yu Li}{
	is a Ph.D. student from the college of computer science, Nankai university.
	He is supervised via Prof. Ming-Ming cheng.
	His research interests include deep learning, machine learning and computer vision.
}

\AuthorBio{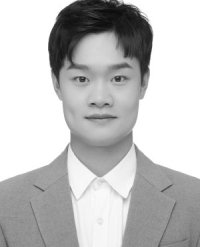}{Qi Han}{
	is a master student from the College of Computer Science, Nankai University,
	under the supervision of Prof. Ming-Ming Cheng.
	He received his bachelor degree from Xidian University in 2019.
	His research interests include deep learning and computer vision.
}

\AuthorBio{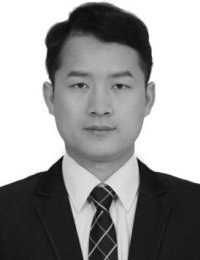}{Ming-Ming Cheng}{
	received his PhD degree from Tsinghua University in 2012,
	and then worked with Prof. Philip Torr in Oxford for 2 years.
	He is now a professor at Nankai University, leading the
	Media Computing Lab.
	His research interests includes computer vision and computer graphics.
	He received awards including ACM China Rising Star Award,
	IBM Global SUR Award, \etc.
	He is a senior member of the IEEE and on the editorial boards of
	IEEE TPAMI and IEEE TIP.
}

\AuthorBio{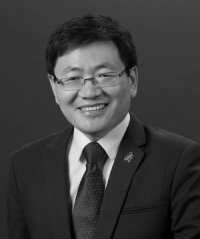}{Liang Wang}{
	received his Ph.D. degree from the Institute of Automation, Chinese Academy of Sciences (CASIA) in 2004. From 2004 to 2010, he has been working at Imperial College London, United Kingdom, Monash University, Australia, the University of Melbourne, Australia, and the University of Bath, United Kingdom, respectively. Currently, he is a full Professor at the National Lab of Pattern Recognition, CASIA. His major research interests include machine learning, pattern recognition and computer vision. He is an associate editor of IEEE Transactions on SMC-B. He is currently an IAPR Fellow and Senior Member of IEEE.
}

\vfill

\end{document}